\documentclass[journal]{IEEEtran}
\newcommand{\myparagraph}[1]{\vspace{0.1em}\noindent\textbf{#1}}
\usepackage[colorlinks,hyperindex,breaklinks]{hyperref}
\newcommand{\ie}{\textit{i}.\textit{e}.}
\newcommand{\eg}{\textit{e}.\textit{g}.}

\usepackage{pifont}
\newcommand{\cmark}{\ding{51}}%
\newcommand{\xmark}{\ding{55}}%
\usepackage{arydshln}

\usepackage{pifont}
\usepackage{colortbl}
\usepackage{arydshln}
\usepackage{multirow}
\usepackage{multicol}
\usepackage{mathrsfs}
\usepackage{bm}
\usepackage{amsfonts}
\usepackage[dvipsnames]{xcolor}
\usepackage{tikz}
\usetikzlibrary{backgrounds}
\usetikzlibrary{arrows,shapes}
\usetikzlibrary{tikzmark}
\usetikzlibrary{calc}
\usepackage{amssymb}
\usepackage{mathtools, nccmath}
\usepackage{wrapfig}
\usepackage{comment}
\usepackage{blindtext}
\usepackage{xspace}
\usepackage{array}
\usepackage{ragged2e}
\newcolumntype{P}[1]{>{\RaggedRight\hspace{0pt}}p{#1}}
\newcolumntype{X}[1]{>{\RaggedRight\hspace*{0pt}}p{#1}}
\usepackage{tcolorbox}
\usepackage{stfloats}
\usepackage{tikz}
\usetikzlibrary{arrows,shapes,positioning,shadows,trees,mindmap}
\usepackage[edges]{forest}
\usetikzlibrary{arrows.meta}
\colorlet{linecol}{black!75}
\usepackage{xkcdcolors}
\newcommand{\highlight}[2]{\colorbox{#1!17}{$\displaystyle #2$}}

\colorlet{mhpurple}{Plum!80}
\renewcommand{\highlight}[2]{\colorbox{#1!17}{#2}}

\usepackage{color}
\usepackage{graphicx}
\usepackage{subfigure}
\usepackage{algorithmic}
\usepackage{mathrsfs}
\usepackage{amsmath}
\usepackage{amsthm}
\usepackage{bm}
\usepackage{algorithm}
\usepackage{float}
\begin{document}
\title{Centralized Feature Pyramid for Object Detection}
\author{Yu~Quan,
        Dong~Zhang,~\IEEEmembership{Member,~IEEE},
        Liyan~Zhang,
        Jinhui~Tang,~\IEEEmembership{Senior Member,~IEEE}
\thanks{Y. Quan, D. Zhang and J. Tang are with the School of Computer Science and Engineering, Nanjing University of Science and Technology, Nanjing 210094, China. E-mail: \{quanyu, dongzhang, jinhuitang\}@njust.edu.cn.

L. Zhang is with the College of Computer Science and Technology, Nanjing University of Aeronautics and Astronautics, MIIT Key Laboratory of Pattern Analysis and Machine Intelligence, Collaborative Innovation Center of Novel Software Technology and Industrialization, Nanjing 211106, China. E-mail: zhangliyan@nuaa.edu.cn. }
\thanks{Corresponding author: Liyan~Zhang.}}
\markboth{under Submission}%
{Shell \MakeLowercase{\textit{et al.}}: Bare Demo of IEEEtran.cls for IEEE Journals}
\maketitle
\begin{abstract}
Visual feature pyramid has shown its superiority in both effectiveness and efficiency in a wide range of applications. However, the existing methods exorbitantly concentrate on the inter-layer feature interactions but ignore the intra-layer feature regulations, which are empirically proved beneficial. Although some methods try to learn a compact intra-layer feature representation with the help of the attention mechanism or the vision transformer, they ignore the neglected corner regions that are important for dense prediction tasks. To address this problem, in this paper, we propose a Centralized Feature Pyramid (CFP) for object detection, which is based on a globally explicit centralized feature regulation. Specifically, we first propose a spatial explicit visual center scheme, where a lightweight MLP is used to capture the globally long-range dependencies and a parallel learnable visual center mechanism is used to capture the local corner regions of the input images. Based on this, we then propose a globally centralized regulation for the commonly-used feature pyramid in a top-down fashion, where the explicit visual center information obtained from the deepest intra-layer feature is used to regulate frontal shallow features. Compared to the existing feature pyramids, CFP not only has the ability to capture the global long-range dependencies, but also efficiently obtain an all-round yet discriminative feature representation. Experimental results on the challenging MS-COCO validate that our proposed CFP can achieve the consistent performance gains on the state-of-the-art YOLOv5 and YOLOX object detection baselines. The code has been released at:~\href{https://github.com/QY1994-0919/CFPNet}{CFPNet}.
\end{abstract}
\IEEEpeerreviewmaketitle
\section{Introduction}
Object detection is one of the most fundamental yet challenging research tasks in the community of computer vision, which aims to predict a unique bounding box for each object of the input image that contains not only the location but also the category information~\cite{zhao2019object}. In the past few years, this task has been extensively developed and applied to a wide range of potential applications, \eg, autonomous driving~\cite{treml2016speeding} and computer-aided diagnosis~\cite{havaei2017brain}.

The successful object detection methods are mainly based on the Convolutional Neural Network (CNN) as the backbone followed with a two-stage (\eg, Fast/Faster R-CNN~\cite{girshick2015fast,ren2015faster}) or single-stage (\eg, SSD~\cite{liu2016ssd} and YOLO~\cite{redmon2016you}) framework. However, due to the uncertainty object sizes, a single feature scale cannot meet requirements of the high-accuracy recognition performance. To this end, methods (\eg, SSD~\cite{liu2016ssd} and FFP~\cite{dollar2014fast}) based on the in-network feature pyramid are proposed and achieve satisfactory results effectively and efficiently. The unified principle behind these methods is to assign region of interest for each object of different size with the appropriate contextual information and enable these objects to be recognized in different feature layers. 

Feature interactions among pixels or objects are important~\cite{vashishth2020interacte}. We consider that effective feature interaction can make image features see wider and obtain richer representations, so that the object detection model can learn an implicit relation (\ie, the favorable co-occurrence features~\cite{zhang2018context,zhang2019co}) between pixels/objects, which has been empirically proved to be beneficial to the visual recognition tasks~\cite{tan2020efficientdet,ghiasi2019fpn,chen2020feature,zhang2020feature,liu2018path,lin2017feature,yin2020disentangled}. For example, FPN~\cite{lin2017feature} proposes a top-down inter-layer feature interaction mechanism, which enables shallow features to obtain the global contextual information and semantic representations of deep features. NAS-FPN~\cite{ghiasi2019fpn} tries to learn the network structure of the feature pyramid part via a network architecture search strategy, and obtains a scalable feature representation. Besides the inter-layer interactions, inspired by the non-local/self-attention mechanism~\cite{wang2018non,vaswani2017attention}, the finer intra-layer interaction methods for spatial feature regulation are also applied to object detection task, \eg, non-local features~\cite{luo2017non} and GCNet~\cite{cao2019gcnet}. Based on the above two interaction mechanisms, FPT~\cite{zhang2020feature} further proposes an inter-layer cross-layer and intra-layer cross-space feature regulation method, and has achieved remarkable performances. 
\begin{figure}[t]
\centering
\includegraphics[width=.48 \textwidth]{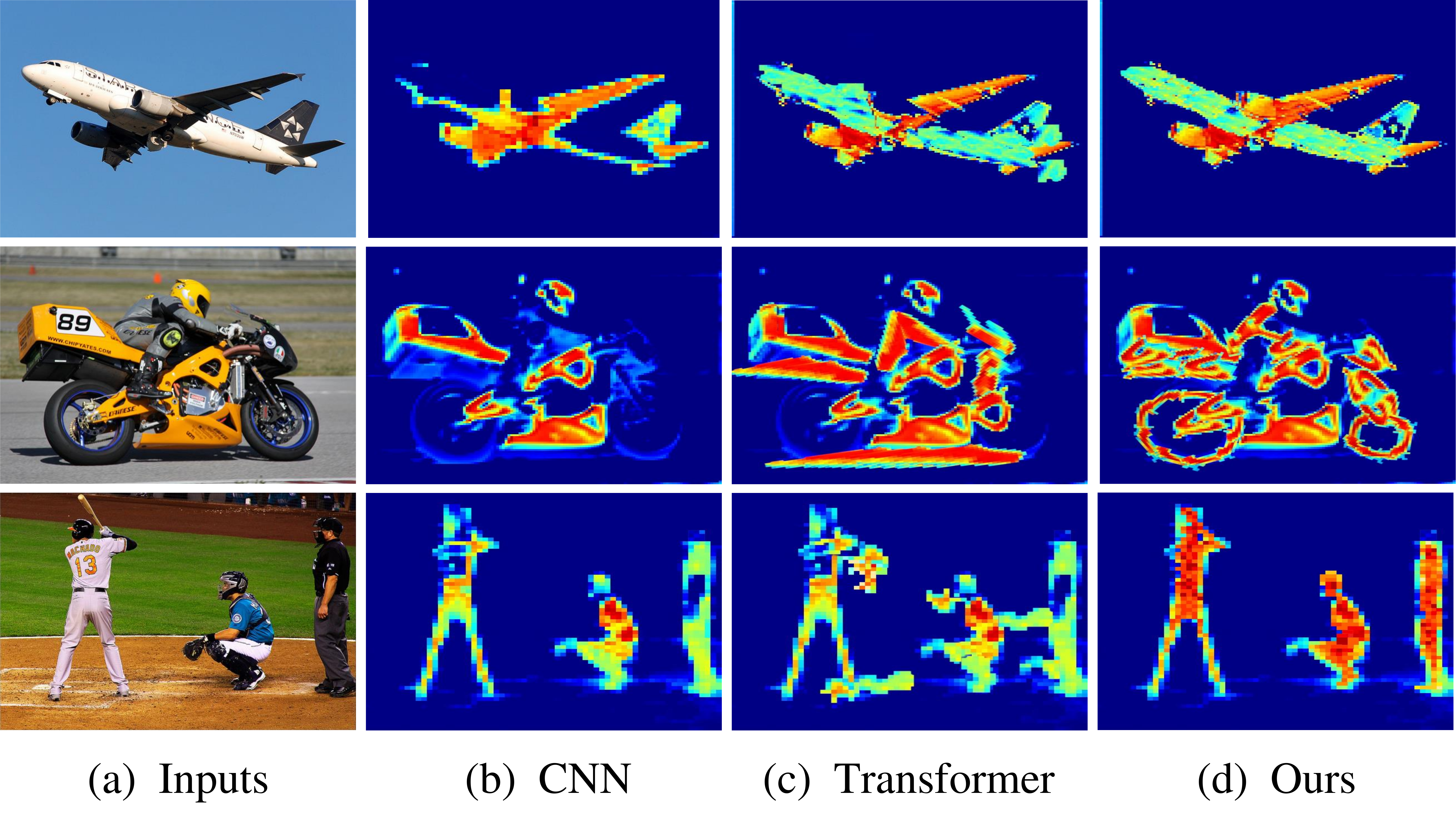}
\caption{Visualizations of image feature evolution for vision recognition tasks. For the input images in (a), a CNN model in (b) only locates those most discriminative regions; although the progressive model in (c) can see wider under the help of the attention mechanism~\cite{vaswani2017attention} or transformer~\cite{carion2020end}, it usually ignores the corner cues that are important for dense prediction tasks; our model in (d) can not only see wider but also more well-rounded by attaching the centralized constraints on features with the advanced long-range dependencies, which is more suitable for dense prediction tasks. Best viewed in color.}
\label{fig1}
\end{figure}

Despite the initiatory success in object detection, the above methods are based on the CNN backbone, which suffer from the inherent limit receptive fields. 
As shown in Figure~\ref{fig1} (a), the standard CNN backbone features can only locate those most discriminative object regions (\eg, the ``body of an  airplane'' and the ``motorcycle pedals''). To solve this problem, vision transformer-based object detection methods~\cite{dosovitskiy2020image,carion2020end,liu2021swin,wang2021pyramid} have been recently proposed and flourished. These methods first divide the input image into different image patches, and then use the multi-head attention-based feature interaction among patches to complete the purpose of obtaining the global long-range dependencies. As expected, the feature pyramid is also employed in a vision transformer, \eg, PVT~\cite{wang2021pyramid} and Swin Transformer~\cite{liu2021swin}. Although these methods can address the limited receptive fields and the local contextual information in CNN, an obvious drawback is their large computational complexity. For example, a Swin-B~\cite{liu2021swin} has almost $3 \times$ model FLOPs (\ie, $47.0$ G vs $16.0$ G) than a performance-comparable CNN model RegNetY~\cite{radosavovic2020designing} with the input size of $224 \times 224$. Besides, as shown in Figure~\ref{fig1} (b), since vision transformer-based methods are implemented in an omnidirectional and unbiased learning pattern, which is easy to ignore some corner regions (\eg, the ``airplane engine'', the ``motorcycle wheel'' and the ``bat'') that are important for dense prediction tasks. These drawbacks are more obvious on the large-scale input images. To this end, we rise a question: is it necessary to use transformer encodes on all layers? To answer such a question, we start from an analysis of shallow features. Researches of the advanced methods~\cite{zhou2016learning,ru2022learning,li2022transcam} show that the shallow features mainly contain some general object feature patterns, \eg, texture, colour and orientation, which are often not global. In contrast, the deep features reflect the object-specific information, which usually requires global information~\cite{strudel2021segmenter,zhu2021unified}. Therefore, we argue that the transformer encoder is unnecessary in all layers.

In this work, we propose a Centralized Feature Pyramid (CFP) network for object detection, which is based on a globally explicit centralized regulation scheme. Specifically, based on an visual feature pyramid extracted from the CNN backbone, we first propose an explicit visual center scheme, where a lightweight MLP architecture is used to capture the long-range dependencies and a parallel learnable visual center mechanism is used to aggregate the local key regions of the input images. Considering the fact that the deepest features usually contain the most abstract feature representations scarce in the shallow features~\cite{zhang2021self}, based on the proposed regulation scheme, we then propose a globally centralized regulation for the extracted feature pyramid in a top-down manner, where the spatial explicit visual center obtained from the deepest features is used to regulate all the frontal shallow features simultaneously. Compared to the existing feature pyramids, as shown in Figure~\ref{fig1} (c), CFP not only has the ability to capture the global long-range dependencies, but also efficiently obtain an all-round yet discriminative feature representation. To demonstrate the superiority, extensive experiments are carried out on the challenging MS-COCO dataset~\cite{lin2014microsoft}. Results validate that our proposed CFP can achieve the consistent performance gains on the state-of-the-art YOLOv5~\cite{yolov5} and YOLOX~\cite{ge2021yolox} object detection baselines.

Our contributions are summarized as the following: 1) We proposed a spatial explicit visual center scheme, which consists of a lightweight MLP for capturing the global long-range dependencies and a learnable visual center for aggregating the local key regions. 2) We proposed a globally centralized regulation for the commonly-used feature pyramid in a top-down manner. 3) CFP achieved the consistent performance gains on the strong object detection baselines.
\section{Related Work}
\subsection{Feature Pyramid in Computer Vision}
Feature pyramid is a fundamental neck network in modern recognition systems that can be effectively and efficiently used to detect objects with different scales. SSD~\cite{liu2016ssd} is one of the first approaches that uses a pyramidal feature hierarchy representation, which captures multi-scale feature information through network of different spatial sizes, thus the model recognition accuracy is improved. FPN~\cite{lin2017feature} hierarchically mainly relies on the bottom-up in-network feature pyramid, which builds a top-down path with lateral connections from multi-scale high-level semantic feature maps. Based on which, PANet~\cite{liu2018path} further proposed an additional bottom-up pathway based on FPN to share feature information between the inter-layer features, such that the high-level features can also obtain sufficient details in low-level features.
Under the help of the neural architecture search, NAS-FPN~\cite{ghiasi2019fpn} uses spatial search strategy to connect across layers via a feature pyramid and obtains the extensible feature information. M2Det~\cite{zhao2019m2det} extracted multi-stage and multi-scale features by constructing multi-stage feature pyramid to achieve cross-level and cross-layer feature fusion. In general, 1) the feature pyramid can deal with the problem of multi-scale change in object recognition without increasing the computational overhead; 2) the extracted features can generate multi-scale feature representations including some high resolution features. In this work, we propose an intra-layer feature regulation from the perspective of inter-layer feature interactions and intra-layer feature regulations of feature pyramids, which makes up for the shortcomings of current methods in this regard.
\subsection{Visual Attention Learning}
CNN~\cite{krizhevsky2012imagenet} focuses more on the representative learning of local regions. However, this local representation does not satisfy the requirement for global context and long-term dependencies of the modern recognition systems. To this end, the attention learning mechanism~\cite{vaswani2017attention} is proposed that focuses on deciding where to project more attention in an image. For example, non-local operation~\cite{wang2018non} uses the non-local neural network to directly capture long-range dependencies, demonstrating the significance of non-local modeling for tasks of video classification, object detection and segmentation. However, the local representation of the internal nature of CNNs is not resolved, \ie, CNN features can only capture limited contextual information. To address this problem, Transformer~\cite{vaswani2017attention} which mainly benefits from the multi-head attention mechanism has caused a great sensation recently and achieved great success in the field of computer vision, such as image recognition~\cite{dosovitskiy2020image,touvron2021training,carion2020end,zhu2020deformable,liu2021swin}. 
For example, the representative VIT divides the image into a sequence with position encoding, and then uses the cascaded transformer block to extract the parameterized vector as visual representations. On this basis, many excellent models~\cite{touvron2021training,beal2020toward,zheng2021rethinking} have been proposed through further improvement, and have achieved good performance in various tasks of computer vision. Nevertheless, the transformer-based image recognition models still have disadvantages of being computationally intensive and complex. 
\subsection{MLP in Computer Vision}
In order to alleviate shortcomings of complex transformer models~\cite{chen20182,ramachandran2019stand,carion2020end,vaswani2021scaling}, recent works~\cite{tolstikhin2021mlp,liu2021pay,yu2021metaformer,hou2022vision} show that replacing attention-based modules in a transformer model with MLP still performs well.
The reason for this phenomenon is that both MLP (\eg, two fully-connected layer network) and attention mechanism are global information processing modules. On the one hand, the introduction of the MLP-Mixer~\cite{tolstikhin2021mlp} into the vision alleviates changes to the data layout. On the other hand, MLP-Mixer can better establish the long dependence/global relationship and spatial relationship of features through the interaction between spatial feature information and channel feature information. Although MLP-style models perform well in computer vision tasks, they are still lacking in capturing fine-grained feature representations and obtaining higher recognition accuracy in object detection. Nevertheless, MLP is playing an increasingly important role in the field of computer vision, and has the advantage of a simpler network structure than transformer. In our work, we also use MLP to capture the global contextual information and long-term dependencies of the input images. Our contribution lies in the centrality of the grasped information using the proposed spatial explicit visual center scheme.
\begin{figure*}[t]
\centering
\includegraphics[width=.9 \textwidth]{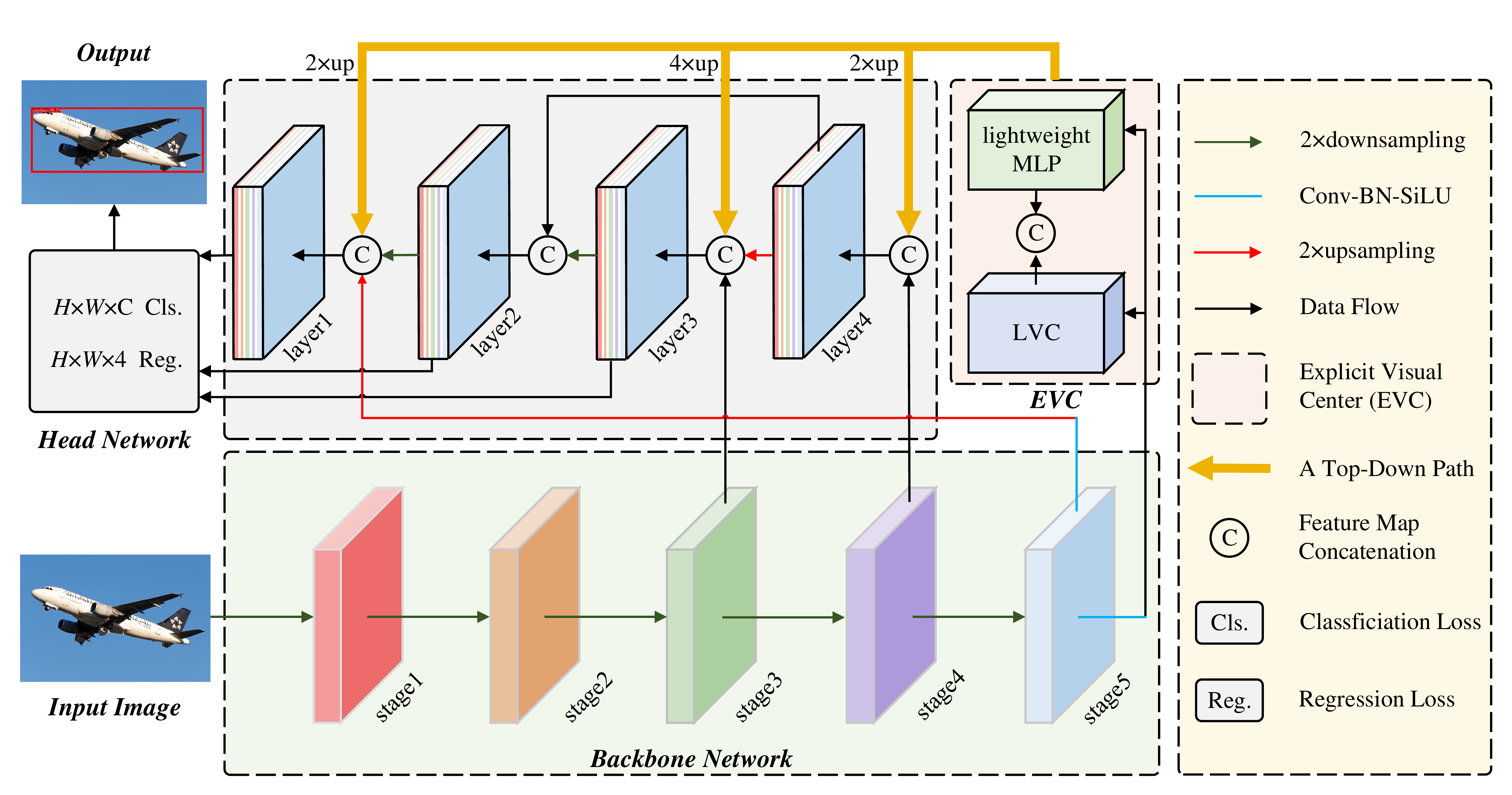}
\vspace{-2mm}
\caption{An illustration of the overall architecture, which mainly consists of four components: input image, a backbone network for feature extraction, the centralized feature pyramid which is based on a commonly-used vision feature pyramid following~\cite{ge2021yolox}, and the object detection head network which includes a classification (\ie, Cls.) loss and a regression (\ie, Reg.) loss. $C$ denotes the class size of the used dataset. Our contribution lines in that we propose an intra-layer feature regulation method in a feature pyramid, and a top-to-down global centralized regulation.}
\vspace{-3mm}
\label{fig2}
\end{figure*}
\subsection{Object Detection}
Object detection is a fundamental computer vision task, which aimes to recognize objects or instances of interest for the given image and provide a comprehensive scene description including the object category and location. With the unprecedented development of CNN~\cite{krizhevsky2012imagenet} in the recent years, plenty of object detection models achieve remarkable progress. 
The existing methods can be divided into two types of two-stage and single-stage. Two-stage object detectors~\cite{girshick2014rich,girshick2015fast,ren2015faster,dai2016r,he2017mask} usually first use a RPN to generate a collection of region proposals. Then use a learning module to extract region features of these region proposals and complete the classification and regression process. 
However, storing and repetitively extracting the features of each region proposal is not only computationally expensive, but also makes it impossible to capture the global feature representations. 
To this end, the single-stage detectors~\cite{redmon2016you,liu2016ssd,redmon2018yolov3,bochkovskiy2020yolov4} directly perform prediction and region classification by generating bounding boxes. The existing single-stage methods have a global concept in the design of feature extraction, and use the backbone network to extract feature maps of the entire image to predict each bounding box. 
In this paper, we also choose the single-stage object detectors (\ie, YOLOv5~\cite{yolov5} and YOLOX~\cite{ge2021yolox}) as our baseline models. Our focus is to enhance the representation of the feature pyramid used for these detectors.

\section{Our Approach}\label{sec:3}
In this section, we introduce the implementation details of the proposed centralized feature pyramid (CFP). We first make an overview architecture description for CFP in Section~\ref{sec:3:1}. Then, we show the implementation details of the explicit visual center in Section~\ref{sec:3:2}. Finally, we show how to implement the explicit visual center on an image feature pyramid and propose our global centralized regulation in Section~\ref{sec:3:3}.

\subsection{Centralized Feature Pyramid (CFP)}\label{sec:3:1}
Although the existing methods have been largely concentrated on the inter-layer feature interactions, they ignore the intra-layer feature regulations, which have been empirically proved beneficial to the vision recognition tasks.
In our work, inspired by the previous works on dense prediction tasks~\cite{peng2021conformer,yu2021metaformer,tolstikhin2021mlp}, we propose a CFP for object detection, which is based on the globally explicit centralized intra-layer feature regulation. Compared to the existing feature pyramids, our proposed CFP not only can capture the global long-range dependencies, but also enable comprehensive and differentiated feature representations. As illustrated in Figure~\ref{fig2}, CFP mainly consists of the following parts: the input image, a CNN backbone is used to extract the vision feature pyramid, the proposed Explicit Visual Center (EVC), the proposed Global Centralized Regulation (GCR), and a decoupled head network (which consists of a classification loss, a regression loss and a segmentation loss) for object detection. In Figure~\ref{fig2}, EVC and GCR are implemented on the extracted feature pyramid. 

Concretely, we first feed the input image into the backbone network (\ie, the Modified CSP v5~\cite{wang2020cspnet}) to extract a five-level one feature pyramid $\textbf{X}$, where the spatial size of each layer of features $\textbf{X}_i$~($i=0, 1, 2, 3, 4$) is $1/2$, $1/4$, $1/8$, $1/16$, $1/32$ of the input image, respectively. Based on this feature pyramid, our CFP is implemented. A lightweight MLP architecture is proposed to capture the global long-range dependencies of $\textbf{X}_4$, where the multi-head self-attention module of a standard transformer encoder is replaced by a MLP layer. Compared to the transformer encoder based on the multi-head attention mechanism, our lightweight MLP architecture is not only simple in structure but also has a lighter volume and higher computational efficiency (cf. Section~\ref{sec:3:2}). Besides, a learnable visual center mechanism, along with the lightweight MLP, is used to aggregate the local corner regions of the input image. We name the above parallel structure network as the spatial EVC, which is implemented on the top layer (\ie, $\textbf{X}_4$) of the feature pyramid. Based on the proposed ECV, to enable the shallow layer features of the feature pyramid to benefit from the visual centralized information of the deepest feature at the same time in an efficient pattern, we then propose a GCR in a top-down fashion, where the explicit visual center information obtained from the deepest intra-layer feature is used to regulate all the frontal shallow features (\ie, $\textbf{X}_3$ to $\textbf{X}_2$) simultaneously. Finally, we aggregate these features into a decoupled head network for classification and regression.
\begin{figure*}[t]
\centering
\includegraphics[width=.9 \textwidth]{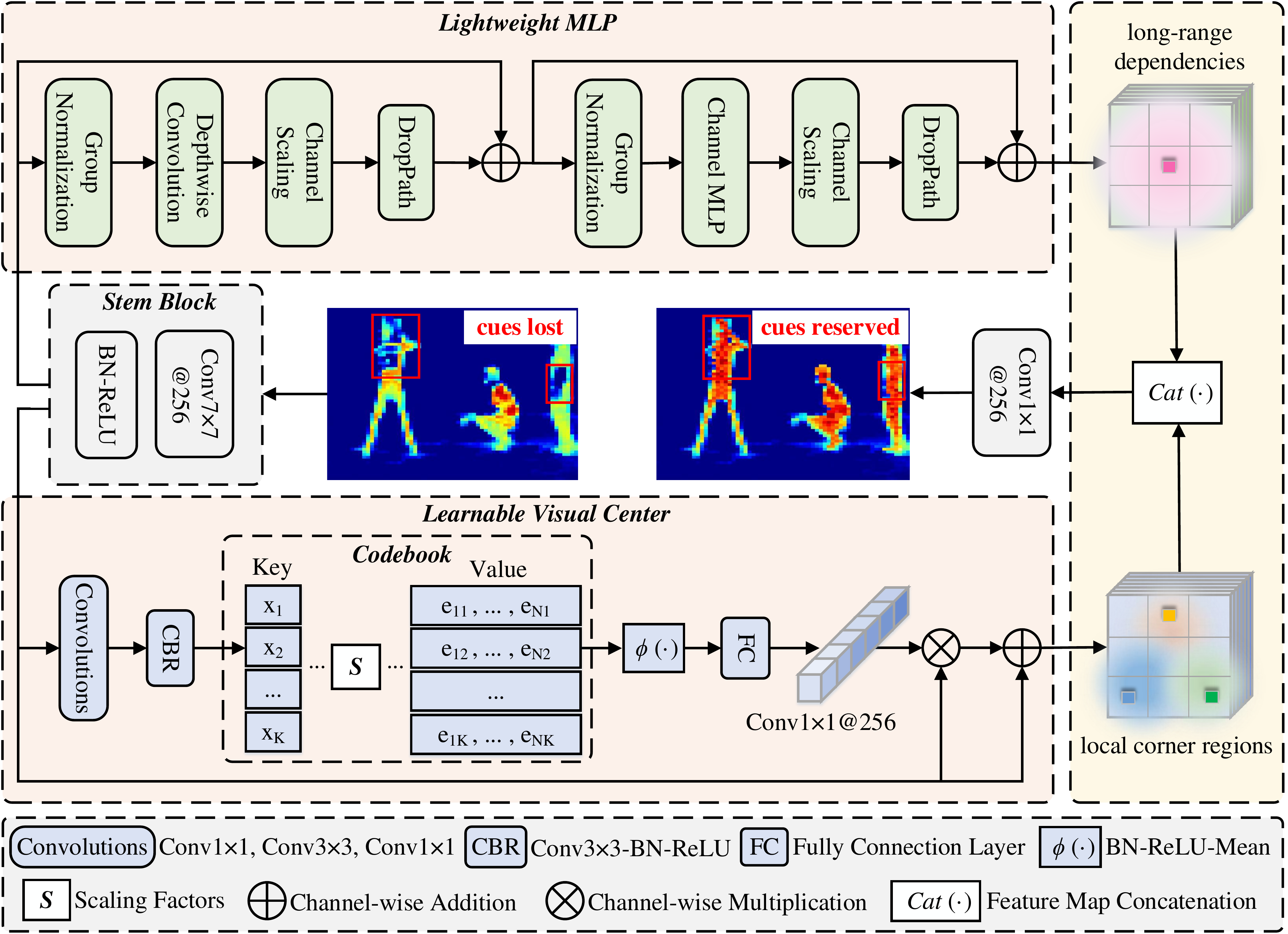}
\caption{An illustration of the proposed explicit visual center, where a lightweight MLP architecture is used to capture the long-range dependencies and a parallel learnable visual center mechanism is used to aggregate the local corner regions of the input image. The integrated features contain advantages of these two blocks, so that the detection model can learn an all-round yet discriminative feature representation.}
\label{fig3}
\end{figure*}
\subsection{Explicit Visual Center (EVC)}\label{sec:3:2}
As illustrated in Figure~\ref{fig3}, our proposed EVC mainly consists of two blocks connected in parallel, where a lightweight MLP is used to capture the global long-range dependencies (\ie, the global information) of the top-level features $\textbf{X}_4$. At the same time, to reserve the local corner regions (\ie, the local information), we propose a learnable vision center mechanism is implemented on $\textbf{X}_4$ to aggregate the intra-layer local regional features. The result feature maps of these two blocks are concatenate together along the channel dimension as the output of EVC for the downstream recognition. In our implementation, between $\textbf{X}_4$ and EVC, a Stem block is used for features smoothing instead of implementing directly on the original feature maps as in~\cite{yolov5}. The Stem block consists of a $7 \times 7$ convolution with the output channel size of $256$, followed by a batch normalization layer, and an activation function layer. The above processes can be formulated as:
\vspace{10mm}
\begin{equation}
\vspace{\baselineskip}
\label{eq2}
\textbf{X} = \textrm{cat}\left(\tikzmarknode{x}{\highlight{red}{$\textrm{MLP}(\textbf{X}_{\textrm{in}})$}}; \tikzmarknode{s}{\highlight{blue}{$\textrm{LVC}(\textbf{X}_{\textrm{in}})$}}\right),
\end{equation}
\begin{tikzpicture}[overlay,remember picture,>=stealth,nodes={align=left,inner ysep=1pt},<-]
\path (x.north) ++ (0,2em) node[anchor=south east,color=red!67] (scalep){\textbf{\small long-range dependencies}};
\draw [color=red!87](x.north) |- ([xshift=-0.3ex,color=red]scalep.south west);
\path (s.south) ++ (0,-.2em) node[anchor=north west,color=blue!67] (mean){\textbf{\small local corner regions}};
\draw [color=blue!87](s.south) |- ([xshift=-0.3ex,color=blue]mean.south east);
\end{tikzpicture}
where $\textbf{X}$ is the output of EVC. $\textrm{cat}(\cdot)$ denotes the feature map concatenation along the channel dimension. ${\highlight{red}{$\textrm{MLP}(\textbf{X}_{\textrm{in}})$}}$ and ${\highlight{blue}{$\textrm{LVC}(\textbf{X}_{\textrm{in}})$}}$ denotes the output features of the used lightweight MLP and the learnable visual center mechanism, respectively. $\textbf{X}_{\textrm{in}}$ is the output of the Stem block, which is obtained by:
\begin{equation}
\textbf{X}_{\textrm{in}} = \sigma(\textrm{BN}(\textrm{Conv}_{7\times7}(\textbf{X}_4))),
\label{eq1}
\end{equation}
where $\textrm{Conv}_{7\times7}(\cdot)$ denotes a $7\times 7$ convolution with stride $1$ and the channel size is set to $256$ in our work following~\cite{lin2017feature}. $\textrm{BN}(\cdot)$ denotes a batch normalization layer and $\sigma (\cdot)$ denotes the $\textrm{ReLU}$ activation function. 

\myparagraph{MLP.} The used lightweight MLP mainly consists of two residual modules: a depthwise convolution-based module~\cite{howard2017mobilenets} and a channel MLP-based block, where the input of the MLP-based module is the output of the depthwise convolution-based~\cite{tolstikhin2021mlp} module. These two blocks are both followed by a channel scaling operation~\cite{yu2021metaformer} and DropPath operation~\cite{larsson2016fractalnet} to improve the feature generalization and robustness ability. Specifically, for the depthwise convolution-based module, features output from the Stem module $\textbf{X}_{\textrm{in}}$ are first fed into a depthwise convolution layer, which have been processed by a group normalization (\ie, feature maps are grouped along the channel dimension). Compared to traditional spatial convolution, depthwise convolution can increase the feature representation ability while reducing the computational costs. Then, channel scaling and droppath are implemented. After that, a residual connection of $\textbf{X}_{\textrm{in}}$ is implemented. The above processes can be formulated as: 
\begin{equation}
\tilde{\textbf{X}}_{\textrm{in}} = \textrm{DConv}(\textrm{GN}(\textbf{X}_{\textrm{in}})) + \textbf{X}_{\textrm{in}},
\label{eq3}
\end{equation}
where $\tilde{\textbf{X}}_{\textrm{in}}$ is the output of the depthwise convolution-based module. $\textrm{GN}(\cdot)$ is the group normalization and $\textrm{DConv}(\cdot)$ is a depthwise convolution~\cite{howard2017mobilenets} with the kernel size of $1 \times 1$. 

For the channel MLP-based module, features output from the depthwise convolution-based module $\tilde{\textbf{X}}_{\textrm{in}}$ are first fed to the a group normalization, and then the channel MLP~\cite{tolstikhin2021mlp} is implemented on these features. Compared to space MLP, channel MLP can not only effectively reduce the computational complexity but also meet the requirements of general vision tasks~\cite{ge2021yolox,bochkovskiy2020yolov4}. After that, channel scaling, droppath, and a residual connection of $\tilde{\textbf{X}}_{\textrm{in}}$ are implemented in sequence. The above processes are expressed as: 
\begin{equation}
{\highlight{red}{$\textrm{MLP}(\textbf{X}_{\textrm{in}})$}} = \textrm{CMLP}(\textrm{GN}(\tilde{\textbf{X}}_{\textrm{in}})) + \tilde{\textbf{X}}_{\textrm{in}},
\label{eq4}
\end{equation}
where $\textrm{CMLP}(\cdot)$ is the channel MLP~\cite{tolstikhin2021mlp}. In our paper, for the presentation convenience, we omit channel scaling and droppath in Eq.~\ref{eq3} and Eq.~\ref{eq4}.

\myparagraph{LVC.} LVC is an encoder with an inherent dictionary and has two components: 1) an inherent codebook: $\textbf{B} = \left\{\textbf{b}_1,\textbf{b}_2,\dots,\textbf{b}_K\right\}$, where $N = H\times W$ is the total spatial number of the input features, where $H$ and $W$ denotes the feature map spatial size in height and width, respectively; 2) a set of scaling factors $\textbf{S} = \left\{\textbf{s}_1,\textbf{s}_2,\dots, \textbf{s}_K\right\}$ for the learnable visual centers. Specifically, features from the Stem block $\textbf{X}_{\textrm{in}}$ are first encoded by a combination of a set of convolution layers (which consist of a $1\times 1$ convolution, a $3\times 3$ convolution, and a $1\times 1$ convolution). Then, the encoded features are processed by a CBR block, which consists of a $3\times 3$ convolution with a BN layer and a ReLU activation function. Through the above steps, the encoded features $\check{\textbf{X}}_{\textrm{in}}$ are entered into the codebook. To this end, we use a set of scaling factor $\textbf{s}$ to sequentially make $\check{\textbf{x}}_{\textrm{i}}$ and $\textbf{b}_\textrm{k}$ map the corresponding position information. The information of the whole image with respect to the $k$-th codeword can be calculated by:
\begin{equation}
\textbf{e}_k = \sum_{i=1}^{N}\frac{e^{-\textbf{s}_{k}\left \| \check{\textbf{x}}_i- \textbf{b}_k \right \|^2}}{\sum_{j=1}^{K}e^{-\textbf{s}_{k}\left \| \check{\textbf{x}}_i- \textbf{b}_k \right \|^2}}(\check{\textbf{x}}_i - \textbf{b}_k),
\label{eq5}
\end{equation}
where $\check{\textbf{x}}_{\textrm{i}}$ is $i$-th pixel point, ${\textbf{b}}_{\textrm{k}}$ is $k$-th learnable visual codeword, and ${\textbf{s}}_{\textrm{k}}$ is $k$-th scaling factor. $\check{\textbf{x}}_{\textrm{i}}-\textbf{b}_\textrm{k}$ is the information about each pixel position relative to a codeword. $K$ is the total number of visual centers. After that, we use $\phi$ to fuse all $\textbf{e}_k$, where $\phi$ contains BN layer with ReLU and mean layer. Based on which, the full information of the whole image with respect to the $K$ codewords is calculated as follows.
\begin{equation}
\textbf{e} = \sum_{k=1}^{K} \phi (\textbf{e}_k).
\label{eq6}
\end{equation}
After obtaining the output of the codebook, we further feed $\textbf{e}$ into a fully connection layer and a $1\times 1$ convolution layer to predict features that highlight key classes. After that, we use the channel-wise multiplication between the input features from Stem block $\textbf{X}_\textrm{in}$ and the scaling factor coefficient $\delta (\cdot)$. The above processes are expressed as:
\begin{equation}
\textbf{Z} = \textbf{X}_{\textrm{in}} \otimes (\delta (\textrm{Conv}_{1\times1}(\textbf{e}))),
\label{eq7}
\end{equation}
where $\textrm{Conv}_{1\times1}$ denotes the $1\times 1$ convolution, and $\delta (\cdot)$ is the sigmoid function. $\otimes$ is channel-wise multiplication.
Finally, we perform a channel-wise addition between features $\textbf{X}_{\textrm{in}}$ output from the Stem block and the local corner region features $\textbf{Z}$, which is formulated as:
\begin{equation}
{\highlight{blue}{$\textrm{LVC}(\textbf{X}_{\textrm{in}})$}} = \textbf{X}_{\textrm{in}} \oplus \textbf{Z},
\label{eq8}
\end{equation}
where $\oplus$ is the channel-wise addition.
\subsection{Global Centralized Regulation (GCR)}\label{sec:3:3}
EVC is a generalized intra-layer feature regulation method that can not only extract global long-range dependencies but also preserve the local corner regional information of the input image as much as possible, which is very important for dense prediction tasks. However, using EVC at every level of the feature pyramid would result in a large computational overheads. To improve the computational efficiency of intra-layer feature regulation, we further propose a GCR for a feature pyramid in a top-down manner.
Specifically, as illustrated in Figure~\ref{fig2}, considering the fact that the deepest features usually contain the most abstract feature representations scarce in the shallow features~\cite{zhang2021self,liu2019simple}, our spatial EVC is first implemented on the top layer (\ie, $\textbf{X}_4$) of the feature pyramid. Then, the obtained features $\textbf{X}$ which includes the spatial explicit visual center is used to regulate all the frontal shallow features (\ie, $\textbf{X}_3$ to $\textbf{X}_2$) simultaneously. In our implementation, on each corresponding low-level features, the features obtained in the deep layer are upsampled to the same spatial scale as the low-level features and then are concatenated along the channel dimension.
Based on which, the concatenated features are downsampled by a $1 \times 1$ convolution into the channel size of $256$ as~\cite{lin2017feature}.
In this way, we are able to explicitly increase the spatial weight of the global representations at each layer of the feature pyramid in the top-down path, such that our CFP can effectively achieve an all-round yet discriminative feature representation.


%

\section{Experiments}
\subsection{Dataset and Evaluation Metrics}\label{sec4:1}
\myparagraph{Dataset.} 
In this work, Microsoft Common Objects in Context (MS-COCO)~\cite{lin2014microsoft} is used to validate the superiority of our proposed CFP. MS-COCO contains $80$ classes of the common scene objects, where the \emph{training} set, \emph{val} set and \emph{test} set contains $118$k, $5$k and $20$k images, respectively. In our experiments, for a fair comparison, all the training images are resized into a fix size of $640 \times640$ as in~\cite{lin2017feature}. For data augmentation, we adopt the commonly used Mosaic~\cite{bochkovskiy2020yolov4} and MixUp~\cite{zhang2017Mixup} in our experiments. 
Mosaic can not only enrich the image data, but also indirectly increase our batch size. MixUp can play a role in increasing the model generalization ability. In particular, following~\cite{ge2021yolox}, our model turns the data augmentation strategy off at the last $15$ epochs in training.

\myparagraph{Evaluation metrics.} 
We mainly follow the commonly used object detection evaluation metric -- Average Precision (AP) in our experiments, which including AP$_\textrm{50}$, AP$_\textrm{75}$, AP$_\textrm{S}$, AP$_\textrm{M}$ and AP$_\textrm{L}$. Besides, to quantitative the model efficiency, GFLOPs, Frame Per Second (FPS), Latency and parameters (Params.) are also used. In particular, following~\cite{ge2021yolox}, Latency and FPS are measured without post-processing for the fair comparison.

\subsection{Implementation Details}\label{sec4:2}
\myparagraph{Baselines.} 
To validate the generality of CFP, we use two state-of-the-art baseline models in our experiments, which are YOLOv5~\cite{yolov5} and YOLOX~\cite{ge2021yolox}. 
In our experiments, we use the end-to-end training strategy and employ their default training and inference settings unless otherwise stated.
\begin{itemize}
\item YOLOv5~\cite{yolov5}. The backbone is a modified cross stage partial network v5~\cite{wang2020cspnet} and DarkNet53~\cite{redmon2018yolov3}, where the modified cross stage partial network v5 is used in the ablation study and the DarkNet53 is used in result comparisons with the state-of-the-art. The neck network is FPN~\cite{lin2017feature}. The object detection head is 
the coupled head network, which contains a classification branch and a regression branch. In YOLOv5, according to the scaling of network depth and width, three different scale networks are generated, they are YOLOv5-Small (YOLOv5-S), YOLOv5-Media (YOLOv5-M), and YOLOv5-Large (YOLOv5-L). 
\item YOLOX~\cite{ge2021yolox}. Compared to YOLOv5, the whole network structure of YOLOX remains unchanged except for the coupled head network. In YOLOv5, object detection head is the decoupled head network.
\end{itemize}

\myparagraph{Backbone.} 
In our experiments, two backbones are used.
\begin{itemize}
\item DarkNet53~\cite{redmon2018yolov3}. DarkNet53 mainly consists of $53$ convolutional layers (basically $1\times 1$ with $3\times 3$ convolutions), which is mainly used for the performance comparisons with state-of-the-art methods in Table~\ref{tab7}.
\item Modified CSPNet v5~\cite{yolov5}. For a fair comparison, we choose YOLOv5 (\ie, the Modified CSPNet v5) as our backbone network. 
The output feature maps is the ones from stage5, which consists of three convolution (Conv, BN and SiLU~\cite{ramachandran2017swish}) operations and a spatial pyramid pooling~\cite{he2015spatial} layer ($5\times 5$, $9\times 9$ and $13\times 13$). 
\end{itemize}

\myparagraph{Comparison methods.}
We consider the use of MLP instead of attention-based, which not only performs well but is computationally less expensive. Therefore, we design a series of MLPs and attention-based variants. Through the ablation study, we choose an optimal variant for our LVC mechanism as well as CFP approach, called lightweight MLP.

\begin{figure*}[t]
\centering
\includegraphics[width=.95 \textwidth]{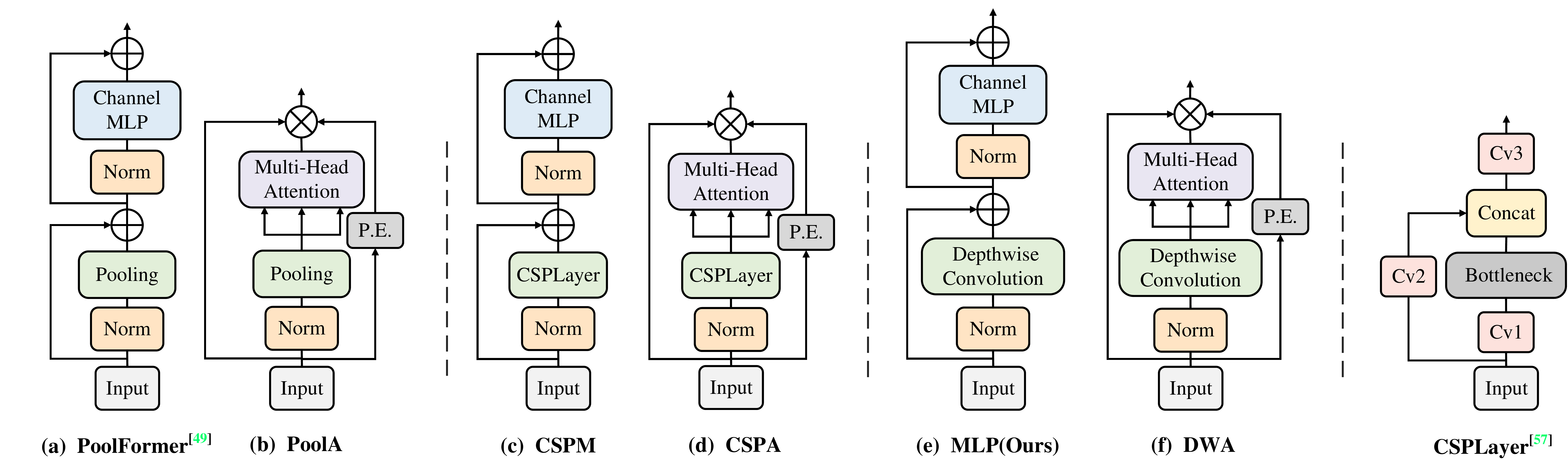}
\caption{MLP and attention-based variants structure diagram. (a) is the PoolFormer structure in~\cite{yu2021metaformer}. (c) and (e) imitate PoolFormer structure and replace Pooling layer with CPSLayer~\cite{wang2020cspnet} and Depthwise Convolution as token mixers respectively. Moreover, (b), (d) and (f) structures replace the channel MLP module with an attention-based module in transformer. Norm denotes the normalization. $\oplus$ represents channel-wise addition operation and $\otimes$ represents channel-wise multiplication operation. P.E. represents positional encoding.}
\label{fig4}
\end{figure*}

Figure~\ref{fig4} (a) shows the PoolFormer structure~\cite{yu2021metaformer}, which consists of a Pooling operation sub-block and a two-layered MLP sub-block.
Considering that the Pooling operation corrupts the detailed features, we choose some convolutions that are structurally lightweight and guarantee accuracy at the same time. Therefore, we designate CPSLayer~\cite{wang2020cspnet} as well as depthwise convolution as token mixers. They are called CSPM and MLP (Ours) in (c) and (e) of Figure~\ref{fig4} respectively. Compared with MLP variants, the structures (b), (d) and (f) are corresponding attention-based variants, respectively. It is worth noting that we choose the channel MLP in the MLP variants. Then, we use convolutional position encoding to prevent absolute position encoding from causing translational invariance of the module. 



\myparagraph{Training settings.} 
We first train our CFP on MS-COCO using pre-trained weights from the YOLOX or YOLOv5 backbone, where all other training parameters are similar in all models. Considering the local hardware condition, our model is trained for a total of $150$ epochs, including $5$ epochs for learning rate warmup as in~\cite{he2016deep}. We use $2$ GeForce RTX $3090$ GPUs with the Batch Size of $16$. Our training settings remain largely consistent from the baseline to final model. The input image training size is $640 \times 640$. The learning rate is set to lr $\times$ BatchSize $/$ $64$ (\ie, the linear scaling strategy~\cite{goyal2017accurate}), where the initial learning rate is set to lr = $0.01$ and the cosine lr schedule is used. The weight decay is set to $0.0005$. The optimizer for the model training process selects stochastic gradient descent, where the momentum is set to $0.9$. Besides, following~\cite{lin2017feature}, we evaluate the AP every $10$ training epochs and report the best one on the MS-COCO \emph{val} set.

\myparagraph{Inference settings.}
For the inference of our model, the original image is scaled to the object size ($640\times 640$) and the rest of the image is filled with gray. Then, we feed the image into the trained model for detection. In the inference that FPS and Latency are all measured with FP16-precision and batch = $1$ on a single GeForce RTX $3090$. However, keep in mind that the inference speed of the models is often uncontrolled, as speed varies with software and hardware.

\subsection{Ablation Study}\label{sec4:3} 
Our ablation study aims to investigate the effectiveness of LVC, MLP, EVC, and CFP in object detection. To this end, we perform a series of experiments on the MS-COCO \emph{val} set~\cite{lin2014microsoft}. From the data in Table~\ref{tab1}, it can be seen that we analyze the effects of LVC, MLP, and EVC on the average precision, the amount of parameters, computation volume, and Latency using YOLOv5-L and YOLOX-L as the baselines, respectively. A detailed analysis of our proposed MLP variants and attention-based variants in terms of precision and Latency is presented in Table~\ref{tab2}, using YOLOX-L as the baseline. Table~\ref{tab3} shows the effect of the number of visual centers $K$ on the LVC at the YOLOX-L baseline. From the data in Table~\ref{tab4}, we can intuitively see the effect of our CFP method on the model with the number of repetitions $R$ at the YOLOX-L baseline.
\begin{table}[t]
\begin{center}
\renewcommand\arraystretch{1.5}
\setlength{\tabcolsep}{1.5pt}{
\caption{Results of ablation study with YOLOv5~\cite{yolov5} and YOLOX-L~\cite{ge2021yolox}.
"$\dagger$" denotes that this is our re-implemented result.}
\resizebox{\linewidth}{!}{
\begin{tabular}{ r | c  c | c  c  c  c } 
Methods & LVC & MLP & mAP ($\%$) & Prams. (M) & GFLOPs (G) & Latency (ms) \\ 
\hline \hline
YOLOv5-L~\cite{yolov5}$\dagger$ & \xmark  & \xmark & 45.20 & 47.10 & 115.60 & 9.34\\
\cdashline{1-7}[0.8pt/2pt]
YOLOv5-L~\cite{yolov5} & \cmark  & \xmark & 46.20$_{\color{red}\textrm{+1.0}}$ & 48.20 & 118.40 & 10.25 \\ 
YOLOv5-L~\cite{yolov5} & \xmark  & \cmark & 45.80$_{\color{red}\textrm{+0.6}}$ & 48.90 & 120.60 & 10.03 \\ 
YOLOv5-L~\cite{yolov5} & \cmark  & \cmark & \textbf{46.60}$_{\color{red}\textrm{+1.4}}$ & 63.90 & 168.00 & 10.68 \\
\hline \hline
YOLOX-L~\cite{ge2021yolox}$\dagger$ & \xmark  & \xmark & 47.80 & 54.21 & 155.65 & 11.01\\
\cdashline{1-7}[0.8pt/2pt]
YOLOX-L~\cite{ge2021yolox} & \cmark  & \xmark &  49.10$_{\color{red}\textrm{+1.3}}$ & 55.31 & 158.30 & 13.09 \\ 
YOLOX-L~\cite{ge2021yolox} & \xmark  & \cmark & 49.10$_{\color{red}\textrm{+1.3}}$ & 56.57 & 163.22 & 12.05 \\ 
YOLOX-L~\cite{ge2021yolox} & \cmark  & \cmark & \textbf{49.20}$_{\color{red}\textrm{+1.4}}$ & 56.83 & 164.05 & 13.36 
\end{tabular}}
\label{tab1}}
\end{center}
\end{table}

\myparagraph{Effectiveness on different baselines.}
As shown in Table~\ref{tab1}, we perform ablation studies on the MS-COCO \emph{val} set using YOLOv5-L and YOLOX-L as baselines for the proposed MLP, LVC, and EVC, respectively. 
As shown in Table~\ref{tab1}, when we only use LVC mechanism to aggregate local corner region features, and the parameters, computation volume and Latency are all within the acceptable growth range, the mAP of our YOLOv5-L and YOLOX-L models are improved by $1.0$\% and $1.3$\%, respectively. Furthermore, when we capture the global long-range dependencies using only the lightweight MLP structure, the mAP of the YOLOv5-L and YOLOX-L models improve by $0.6$\% and $1.3$\%, respectively. 
Most importantly, when we use both LVC and MLP (the EVC scheme) on the YOLOv5-L and YOLOX-L baselines, the mAP of both models are improved by $1.4$\%. Further analysis shows that when EVC scheme is applied to YOLOv5-L baseline and YOLOX-L baseline respectively, mAP of YOLOX-L model can be improved to $49.2$\%, and its parameter number and computation volume are lower than those of YOLOv5-L model. 
The results show that the EVC scheme is more effective in YOLOX-L baseline, and the overhead is slightly smaller than that of YOLOv5-L baseline. YOLOX-L is used as the baseline in subsequent ablation experiments.

\begin{table*}[t]
\begin{center}
\renewcommand\arraystretch{1.5}
\setlength{\tabcolsep}{11pt}{
\caption{Result comparisons of our lightweight MLP(Ours)$_{\textrm{YOLOX-L}}$ with the MLP variants and the self-attention variants on the MS-COCO \emph{val} set~\cite{lin2014microsoft}. "$\dagger$" is our re-implementation result.}
\begin{tabular}{ r | c  c  c  c  c  c | c} 
Methods & mAP ($\%$) & AP$_{50}$ ($\%$) & AP$_{75}$ ($\%$) & AP$_S$ ($\%$) & AP$_M$ ($\%$) & AP$_L$ ($\%$)  & Latency (ms) \\ 
\hline \hline
YOLOX-L~\cite{ge2021yolox}$\dagger$ & 47.80  & 65.80 & 51.60 & 29.90 & 52.80 & 62.40 & 11.01 \\
\cdashline{1-8}[0.8pt/2pt]
PoolFormer$_{\textrm{YOLOX-L}}$~\cite{yu2021metaformer}$\dagger$ & 47.80$_{\color{red}\textrm{+0.0}}$ & 65.70 & 51.70 & 29.90 & 52.90 & 62.40 & 11.79\\
PoolA$_{\textrm{YOLOX-L}}$ &  47.90$_{\color{red}\textrm{+0.1}}$ & 65.90 & 51.80 & 30.10 & 52.70 & 62.50 & 12.03 \\ 
CSPM$_{\textrm{YOLOX-L}}$ & 47.70$_{\color{blue}\textrm{-0.1}}$ & 65.70 & 51.50 & 29.50 & 52.50 & 62.80 & 11.75\\ 
CSPA$_{\textrm{YOLOX-L}}$ & 48.00$_{\color{red}\textrm{+0.2}}$ & 66.00 & 52.10 & 30.50 & 53.0 & 62.60 & 12.11\\ 
DWA$_{\textrm{YOLOX-L}}$ & 49.20$_{\color{red}\textrm{+1.4}}$ & 67.70 & 53.20 & 31.60 & 54.00 & 63.90 & 14.89\\ 
\cdashline{1-8}[0.8pt/2pt]
MLP$_{\textrm{YOLOX-L}}$ (Ours) & \textbf{49.10}$_{\color{red}\textrm{+1.3}}$ & 67.50 & 53.30 & 31.20 & 53.70 & 63.90 & 12.05
\end{tabular}
\label{tab2}}
\end{center}
\end{table*}

\myparagraph{Comparisons with MLP variants.} 
Table~\ref{tab2} shows the detection performance of MLP and attention-based variants based on YOLOX-L baseline on the MS-COCO \emph{val} set. We first analyze the comparison results of MLP variants. 
We can observe that the PoolFormer structure obtains the same mAP (\ie, $47.80$\%) as YOLOX-L model. 
The performance of CSPM is even worse compared to the YOLOX-L model, which not only reduces the average precision by $0.1$\%, but also increases the Latency by $0.74$ms. 
But our proposed lightweight MLP structure obtains the highest mAP (\ie, $49.10$\%) in the MLP variants, which is $1.3$\% better than the mAP of YOLOX-L. This also demonstrates that our choice of depthwise convolution as the token mixer in the MLP variant performs better. 
Turning to the attention-based variants, the performance of PoolA, CSPA and DWA are all improved compared to YOLOX-L, and the mAP of DWA can reach $49.20$\%. But in fact, we compare the two best performing structures (MLP and DWA) find that the Latency of DWA increases by $2.84$ms in the same hardware environment than MLP (Ours). 
From the comprehensive analysis of the data in Table~\ref{tab2}, it can be found that our lightweight MLP is not only better but also faster in capturing long-range dependencies.

\myparagraph{Effect of $K$.} 
As shown in Table~\ref{tab3}, we analyze the effect of the number of visual centers $K$ on the performance of LVC. We choose YOLOX-L as the baseline, and with increasing $K$, we can observe that its performance shows an increasing trend. At the same time, the parameter number, computation volume and the Latency of the model also tend to increase gradually. Notably, when $K$ = $64$, the mAP of the model can reach $49.10$\% and when $K$ = $128$, the mAP of the model can reach $49.20$\%. 
Although the performance of the model improves by $0.1$\% as $K$ increases, its extra computational cost increases by 10.01G, and the corresponding inference time increases by 3.21ms. The reason for this may be that too much visual centers bring more redundant semantic information. Not only the performance is not significantly improved, but also the computational effort is increased. So we choose $K$ = $64$. 
\begin{table}[t]
\begin{center}
\renewcommand\arraystretch{1.5}
\setlength{\tabcolsep}{1.5pt}{
\caption{Effect of the number of visual centers $K$ on LVC$_{\textrm{YOLOX-L}}$. ``-'' denotes that there is no such a setting. "$\dagger$" is our re-implementation result.}
\resizebox{\linewidth}{!}{
\begin{tabular}{ r | c | c c c c } 
Methods & $K$ & mAP ($\%$) & Params. (M) & GFLOPs (G) & Latency (ms) \\ 
\hline \hline
YOLOX-L~\cite{ge2021yolox}$\dagger$  & - & 47.80 & 54.21 & 155.65 & 11.01 \\  
\cdashline{1-6}[0.8pt/2pt]
LVC$_{\textrm{YOLOX-L}}$ & 16 & 47.90$_{\color{red}\textrm{+0.1}}$ & 54.21 & 155.65 & 10.99 \\ 
LVC$_{\textrm{YOLOX-L}}$ & 32 & 48.00$_{\color{red}\textrm{+0.2}}$ & 55.28 & 158.29 & 12.24 \\ 
LVC$_{\textrm{YOLOX-L}}$ & 64 & \textbf{49.10}$_{\color{red}\textrm{+1.3}}$ & 55.31 & 158.30 & 13.09 \\  
LVC$_{\textrm{YOLOX-L}}$ & 128 & 49.20$_{\color{red}\textrm{+1.4}}$ & 57.64 & 168.31 & 16.30 \\  
\end{tabular}}
\label{tab3}}
\end{center}
\end{table}


\myparagraph{Effect of $R$.} 
From Table~\ref{tab4}, we analyze the effect of the number of repetitions $R$ of CFP on the performance. We still choose the YOLOX-L baseline, and as $R$ increases, we can observe a trend of increasing and then decreasing and then stabilizing the performance compared to the YOLOX-L model. Meanwhile, the number of parameters, computation volume, and the Latency all show a gradual increase. 
In particular, when $R$ = $1$, CFP$_{\textrm{YOLOX-L}}$ achieves the best performance mAP of $49.40$\%. When $R$ = $2$, the performance is instead reduced by $0.2$\% compared to $R$ = $1$. The reason may be that this repeated extraction of features does not capture useful information except for increasing the computational cost. Therefore, based on the above observations, we choose $R$ = $1$.
\begin{table}[t]
\begin{center}
\renewcommand\arraystretch{1.5}
\setlength{\tabcolsep}{1.5pt}{
\caption{Result comparisons of the number of repetitions of the proposed CFP in the YOLOX-L baseline. ``-'' denotes that there is no such a setting. "$\dagger$" is our re-implementation result. $R$ denotes the number of repetitions.}
\resizebox{\linewidth}{!}{
\begin{tabular}{ r | c | c c c c } 
Methods & $R$ & mAP ($\%$) & Params. (M) & GFLOPs (G) & Latency (ms) \\ 
\hline \hline
YOLOX-L~\cite{ge2021yolox}$\dagger$  & - & 47.80 & 54.21 & 155.65 & 11.01 \\  
\cdashline{1-6}[0.8pt/2pt]
CFP$_{\textrm{YOLOX-L}}$ & 1 & \textbf{49.40}$_{\color{red}\textrm{+1.6}}$ & 71.14 & 209.07 & 15.45 \\ 
CFP$_{\textrm{YOLOX-L}}$ & 2 & 49.20$_{\color{red}\textrm{+1.4}}$ & 72.09 & 211.78 & 15.56 \\ 
CFP$_{\textrm{YOLOX-L}}$ & 3 & 49.20$_{\color{red}\textrm{+1.4}}$ & 74.56 &  234.28 & 16.37 \\  
\end{tabular}}
\label{tab4}}
\end{center}
\end{table}

\subsection{Efficiency Analysis}\label{sec4:4}
\begin{figure}[t]
\centering
\includegraphics[width=.48 \textwidth]{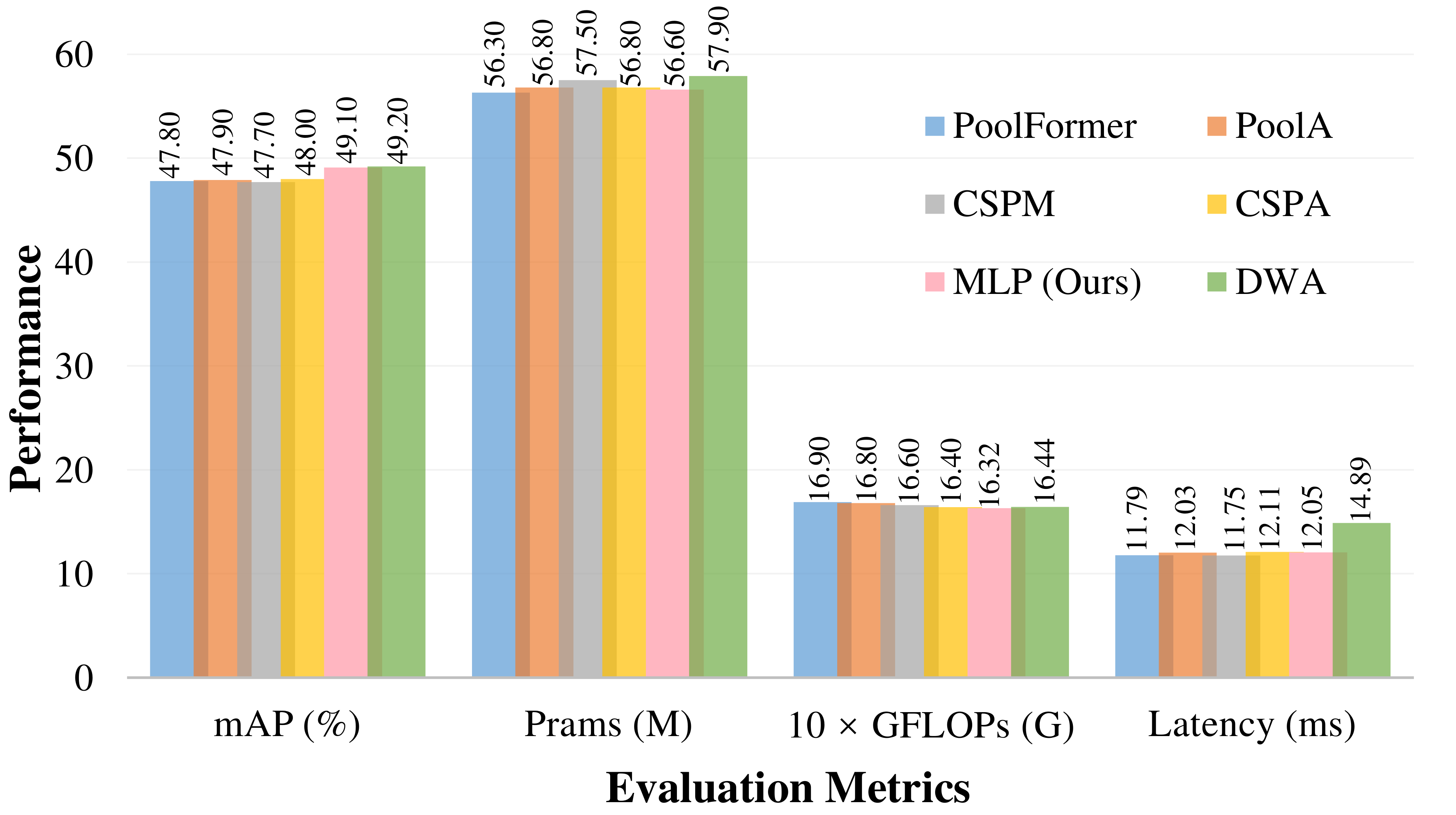}
\caption{Multi-metrics comparison results between MLP variants and attention-based variants based on the MS-COCO \emph{val} set.}
\label{fig5}
\end{figure}

In this section, we show the efficiency analysis. 
First, we analyze the performance of the MLP variants and attention-based variants from a multi-metric perspective. In Figure~\ref{fig5}, all models take YOLOX-L as baseline and are trained on the MS-COCO emph{val} set with the same data augmentation settings.
Meanwhile, to demonstrate the effectiveness of the MLP structure, as shown in Table~\ref{tab5}, we compare it with the state-of-the-art transformer methods and the MLP methods at this stage.
As can be observed from Figure~\ref{fig5}, we can intuitively see that the MLP (Ours) structure is significantly better than the other structures in terms of mAP, and it is lower than the other structures in terms of number of parameters, computation volume, and the inference time. 
It can be shown that the MLP structure can guarantee a lower number of parameters and computation volume under the condition of obtaining a better precision. 

In Table~\ref{tab5}, we give the comparative results of the MLP and transformer methods that are outstanding performers in object detection tasks at this stage. IN the first half of Table~\ref{tab5}, our MLP$_{\textrm{YOLOX-L}}$ method not only occupies less memory but also has an average precision of $1.3$\% higher compared to Mask R-CNN (backbone as AS-MLP-S~\cite{lian2021mlp}).
In the middle part of Table~\ref{tab5}, we can find that our MLP$_{\textrm{YOLOX-L}}$ can improve the mAP by up to $7.1$\% compared to the transformer method (DETR~\cite{carion2020end}) without extra computational cost. With the same mAP, the number of parameters of MLP is reduced by $62.43$M compared to REGO-Deformable DETR~\cite{chen2021recurrent}. 
Therefore, we can find that MLP not only has high precision but also takes up less memory compared to the transformer methods. All in all, our MLP has outstanding performance in capturing feature long-range dependencies.
\begin{table*}[t]
\begin{center}
\renewcommand\arraystretch{1.5}
\setlength{\tabcolsep}{18pt}{
\caption{Result comparisons of our lightweight MLP with transformer variants and MLP variant methods. Model efficiency analysis. ``-'' denotes that there is no such a setting. "$\dagger$" is our re-implementation result.}
\begin{tabular}{ r | r | c c c } 
Methods & Backbone & mAP ($\%$) & Params. (M) & GFLOPs (G)\\ 
\hline \hline
MetaFormer~\cite{yu2021metaformer} & PoolFormer-S12~\cite{yu2021metaformer} & 37.30 & 31.60 & 62.75 \\ 
MetaFormer~\cite{yu2021metaformer} & PoolFormer-S24~\cite{yu2021metaformer} & 40.10 & 41.00 & 66.38 \\ 
MetaFormer~\cite{yu2021metaformer} & PoolFormer-S36~\cite{yu2021metaformer} & 41.00 & 50.50 & 67.89 \\
Mask R-CNN~\cite{lian2021mlp} & AS-MLP-T~\cite{lian2021mlp} & 46.00 & 48.00 & 117.45  \\ 
Mask R-CNN~\cite{lian2021mlp} & AS-MLP-S~\cite{lian2021mlp} & 47.80 & 69.00 & 176.20 \\ 
\cdashline{1-5}[0.8pt/2pt]
DETR~\cite{carion2020end} & ResNet-50~\cite{lin2017feature} & 42.00 & 41.00 & 86.00 \\ 
Deformable DETR~\cite{zhu2020deformable} & ResNet-50~\cite{lin2017feature} & 43.80 & 40.00 & 173.00 \\ 
REGO-Deformable DETR~\cite{chen2021recurrent} & ResNetXt-101~\cite{xie2017aggregated} & 49.10 & 119.00 & 434.00 \\ 
YOLOS~\cite{fang2021you} & DeiT-base~\cite{touvron2021training} & 42.00 & 127.00 & 567.00  \\ 
ViDT (w.o. Neck)~\cite{song2021vidt} & Swin-base~\cite{liu2021swin} & 43.20 & 91.00 & 203.40 \\
\cdashline{1-5}[0.8pt/2pt]
MLP$_{\textrm{YOLOv5-L}}$ (Ours) & Modified CSP v5~\cite{yolov5} & \textbf{45.80} & 48.90 & 120.60 \\
MLP$_{\textrm{YOLOX-L}}$ (Ours) & Modified CSP v5~\cite{yolov5} & \textbf{49.10} & 56.57 & 163.22
\end{tabular}
\label{tab5}}
\end{center}
\end{table*}

\subsection{Comparisons with State-of-the-art Methods.}\label{sec4:5} 
\begin{figure*}[t]
\centering
\includegraphics[width=.9 \textwidth]{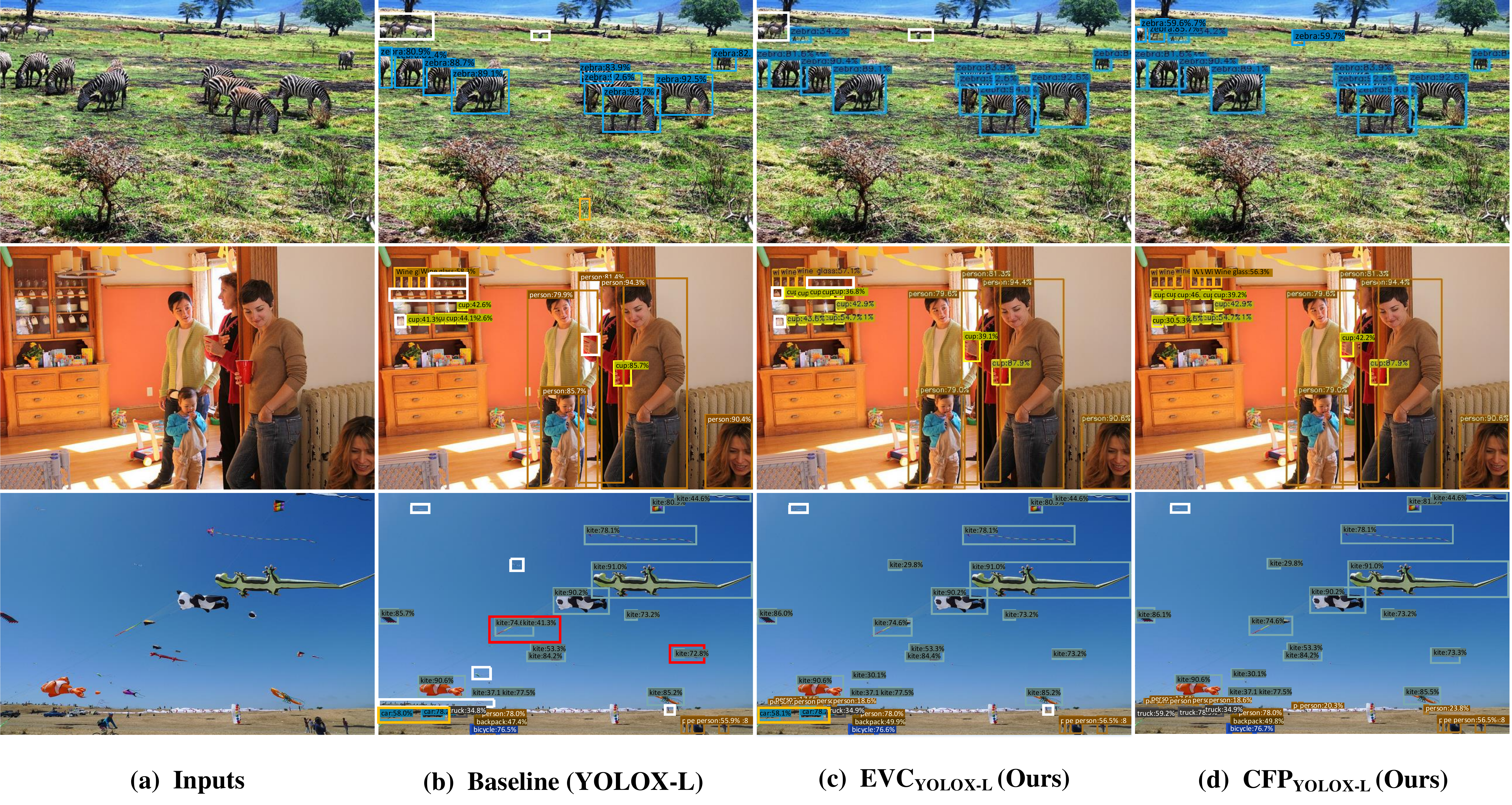}
\caption{Qualitative results on the \emph{test} set of  MS-COCO $2017$~\cite{lin2014microsoft}. We show the results of object detection from baseline and our approaches for comparison.}
\label{fig6}
\end{figure*}

As shown in Table~\ref{tab6}, we validate the proposed CFP method on the MS-COCO \emph{val} set with YOLOv5 (Small, Media and Large) and YOLOX (Small, Media and Large) as baselines. In addition, the data in table~\ref{tab7} show the comparison results of our CFP method compared to the advanced single-stage and two-stage detectors.
Finally, we show some visual comparison plots in Figure~\ref{fig6}.

\begin{table*}[t]
\begin{center}
\renewcommand\arraystretch{1.5}
\setlength{\tabcolsep}{11pt}{
\caption{Result comparisons with YOLOv5 and YOLOX baselines. "$\dagger$" is our re-implementation result.}
\begin{tabular}{ r | r |c  c  c  c  c  c } 
Methods & Backbone & mAP ($\%$) & AP$_{50}$ ($\%$) & AP$_{75}$ ($\%$) & AP$_S$ ($\%$) & AP$_M$ ($\%$) & AP$_L$ ($\%$) \\ 
\hline \hline
YOLOv5~\cite{yolov5}$\dagger$ & Small~\cite{yolov5} & 35.50  & 55.30 & 47.10 & - & - & -  \\ 
YOLOv5~\cite{yolov5}$\dagger$ & Media~\cite{yolov5} & 42.70  & 62.30 & 48.00 & - & - & -  \\ 
YOLOv5~\cite{yolov5}$\dagger$ & Large~\cite{yolov5} & 45.20  & 64.10 & 49.20 & - & - & -  \\ 
\cdashline{1-8}[0.8pt/2pt]
CFP$_{\textrm{YOLOv5}}$~\cite{yolov5} & Small~\cite{yolov5} & \textbf{36.00}$_{\color{red}\textrm{+0.5}}$ & 56.20 & 47.80 & 22.80 & 42.90 & 51.60  \\ 
CFP$_{\textrm{YOLOv5}}$~\cite{yolov5} & Media~\cite{yolov5} & \textbf{43.20}$_{\color{red}\textrm{+0.5}}$ & 62.90 & 48.50 & 29.10 & 49.40 & 53.30  \\ 
CFP$_{\textrm{YOLOv5}}$~\cite{yolov5} & Large~\cite{yolov5} & \textbf{46.60}$_{\color{red}\textrm{+1.4}}$ & 64.90 & 50.00 & 30.40 & 51.70 & 59.50  \\ 
\hline \hline
YOLOX~\cite{ge2021yolox}$\dagger$ & Small~\cite{ge2021yolox} & 34.10  & 52.00 & 36.90 & 18.80 & 38.10 & 44.40  \\ 
YOLOX~\cite{ge2021yolox}$\dagger$ & Media~\cite{ge2021yolox} & 45.60  & 64.30 & 48.90 & 28.00 & 50.20 & 59.70  \\ 
YOLOX~\cite{ge2021yolox}$\dagger$ & Large~\cite{ge2021yolox} & 47.80  & 65.80 & 51.60 & 29.90 & 52.80 & 62.40  \\ 
\cdashline{1-8}[0.8pt/2pt]
CFP$_{\textrm{YOLOX}}$~\cite{ge2021yolox} & Small~\cite{ge2021yolox} & \textbf{41.10}$_{\color{red}\textrm{+7.0}}$  & 60.00 & 44.50 & 24.20 & 45.40 & 54.50  \\ 
CFP$_{\textrm{YOLOX}}$~\cite{ge2021yolox} & Media~\cite{ge2021yolox} & \textbf{46.40}$_{\color{red}\textrm{+0.8}}$  & 65.10 & 50.30 & 29.40 & 51.20 & 60.50  \\ 
CFP$_{\textrm{YOLOX}}$~\cite{ge2021yolox} & Large~\cite{ge2021yolox} & \textbf{49.40}$_{\color{red}\textrm{+1.6}}$  & 67.90 & 53.40 & 31.50 & 54.80 & 64.20 
\end{tabular}
\label{tab6}}
\end{center}
\end{table*}

\myparagraph{Comparison with YOLOv5 and YOLOX baseline.}
As shown in Table~\ref{tab6}, when YOLOv5 is chosen as the baseline, the mAP of our CFP method is enhanced by $0.5$\%, $0.5$\%, and $1.4$\% on the Small, Media, and Large size models, respectively. When YOLOX~\cite{ge2021yolox} is used as the baseline, the mAP improves by $7.0$\%, $0.8$\%, and $1.6$\% on the backbone networks of different sizes. 
It is worth noting that the main reason why we choose YOLOv5 (anchor mechanism) and YOLOX (anchor-free mechanism) as the baseline is that the reciprocity of these two models in terms of network structure can fully demonstrate the effectiveness of our CPF approach. 
Most importantly, our CFP method do not perform poorly due to the shortcomings of the YOLOv5 model, which achieve a maximum mAP of $46.60$\%. Meanwhile, our mAP reaches $49.40$\% at YOLOX baseline. Moreover, CFP$_{\textrm{YOLOX}}$ on the small backbone network is improved by $7.0$\% over YOLOX~\cite{ge2021yolox}. 
The main reason for this is that the LVC in our CFP can enhance the feature representations of the local corner regions through visual centers at the pixel-level.

\begin{table*}[t]
\begin{center}
\renewcommand\arraystretch{1.5}
\setlength{\tabcolsep}{1.5pt}{
\caption{Comparison of the speed and accuracy of different object detectors on MS-COCO \emph{val} set. We select all the models trained on 150 epochs for fair comparison. . "$\dagger$" is our re-implementation result.}
\begin{tabular}{ r | c | r | c | c  c  c  c  c  c} 
Methods & Publication & Backbone & FPS (F/s) &mAP ($\%$) & AP$_{50}$ ($\%$) & AP$_{75}$ ($\%$) & AP$_S$ ($\%$) & AP$_M$ ($\%$) & AP$_L$ ($\%$)  \\ 
\hline \hline
Faster R-CNN+++~\cite{he2016deep} & CVPR 2016 & ResNet-101-C4~\cite{xie2017aggregated} & - & 34.90 & 55.70 & 37.40 & 15.60 & 38.70 & 50.90  \\
Faster R-CNN w FPN~\cite{lin2017feature} & CVPR 2017 & ResNet-101-FPN~\cite{xie2017aggregated} & - & 36.20 & 59.10 & 39.00 & 18.20 & 39.00 & 48.20  \\
Mask R-CNN~\cite{he2017mask} & ICCV 2017 & ResNet-50~\cite{lin2017feature} & - & 38.20 & - & - & 21.90 & 40.90 & 49.50  \\
D2Det~\cite{cao2020d2det} & CVPR 2020 & ResNet101-deform v2~\cite{cao2020d2det} & - & 47.40 & 65.90 & 51.70 & 27.20 & 50.40 & 61.30  \\
\cdashline{1-10}[0.8pt/2pt]
SSD513~\cite{liu2016ssd} & ECCV 2016 & ResNet-101-SSD~\cite{liu2016ssd} & - & 21.60 & 44.00 & 19.20 & 5.00 & 22.40 & 35.50  \\
YOLOv2~\cite{redmon2017yolo9000} & CVPR 2017 & DarkNet-19~\cite{redmon2017yolo9000} & - & 21.60 & 44.00 & 19.20 & 5.00 & 22.40 & 35.50  \\
DSSD513~\cite{fu2017dssd} & arXiv 2017 & ResNet-101-DSSD~\cite{fu2017dssd} & - & 31.20 & 50.40 & 33.30 & 10.20 & 34.50 & 49.80 \\
YOLOv3-ultralytics~\cite{redmon2018yolov3} & arXiv 2018 & DarkNet-53~\cite{redmon2018yolov3} & 93.60 & 33.00 & 57.90 & 34.40 & 18.30 & 35.40 & 41.90  \\ 
\cdashline{1-10}[0.8pt/2pt]
EfficientDet-D0~\cite{Tan_2020_CVPR} & CVPR 2020 & EfficientNet-B0~\cite{tan2019efficientnet} & 97.00 & 34.60 & 53.00 & 37.10 & 12.40 & 39.00 & 52.70  \\
EfficientDet-D1~\cite{Tan_2020_CVPR} & CVPR 2020 & EfficientNet-B1~\cite{tan2019efficientnet} & 74.00 & 40.50 & 59.10 & 43.70 & 18.30 & 45.00 & 57.50  \\
EfficientDet-D2~\cite{Tan_2020_CVPR} & CVPR 2020 & EfficientNet-B2~\cite{tan2019efficientnet} & 57.00 & 43.00 & 62.30 & 46.20 & 22.50 & 48.00 & 58.40  \\
EfficientDet-D3~\cite{Tan_2020_CVPR} & CVPR 2020 & EfficientNet-B2~\cite{tan2019efficientnet} & 36.00 & 47.50 & 66.20 & 51.50 & 27.90 & 51.40 & 62.00  \\
\cdashline{1-10}[0.8pt/2pt]
YOLOv4~\cite{bochkovskiy2020yolov4}$\dagger$ & arXiv 2020 & CSPDarkNet-53~\cite{bochkovskiy2020yolov4} & 60.60 & 42.30 & 64.70 & 46.20 & 25.90 & 45.40 & 52.20  \\
YOLOv4-CSP~\cite{wang2021scaled}$\dagger$ & CVPR 2021 & Modified CSP~\cite{wang2021scaled} & 71.90 & 46.70 & 65.30 & 50.20 & 27.20 & 49.90 & 58.70  \\
\cdashline{1-10}[0.8pt/2pt]
YOLOv5-S~\cite{yolov5}$\dagger$ & GitHub 2021 & Modified CSP v5~\cite{yolov5} & 90.40 & 35.50 & 55.30 & 47.10 & - & - & -  \\ 
YOLOv5-M~\cite{yolov5}$\dagger$ & GitHub 2021 & Modified CSP v5~\cite{yolov5} & 89.20 & 42.70 & 62.30 & 48.00 & - & - & -  \\ 
YOLOv5-L~\cite{yolov5}$\dagger$ & GitHub 2021 & Modified CSP v5~\cite{yolov5} & 71.90 & 45.20 & 64.10 & 49.20 & - & - & -  \\ 
\cdashline{1-10}[0.8pt/2pt]
YOLOX-DarkNet53~\cite{ge2021yolox}$\dagger$ & arXiv 2021 & DarkNet-53~\cite{redmon2018yolov3} & 89.20 & 46.30 & 64.60 & 49.10 & 28.40 & 50.80 & 60.70  \\ 
YOLOX-S~\cite{ge2021yolox}$\dagger$ & arXiv 2021 & Modified CSP v5~\cite{yolov5} & 89.90 & 34.10 & 52.00 & 36.90 & 18.80 & 38.10 & 44.40  \\ 
YOLOX-M~\cite{ge2021yolox}$\dagger$ & arXiv 2021 & Modified CSP v5~\cite{yolov5} & 78.70 & 45.60 & 64.30 & 48.90 & 28.00 & 50.20 & 59.70  \\ 
YOLOX-L~\cite{ge2021yolox}$\dagger$ & arXiv 2021 & Modified CSP v5~\cite{yolov5} & 68.00 & 47.80 & 65.80 & 51.60 & 29.90 & 52.80 & 62.40  \\ 
\cdashline{1-10}[0.8pt/2pt]
CFP$_{\textrm{YOLOv5}}$~\cite{yolov5} & None & DarkNet-53~\cite{redmon2018yolov3} & 86.50 & 46.40 & 63.70 & 48.70 & 27.80 & 49.20 & 58.40  \\ 
CFP$_{\textrm{YOLOv5-S}}$~\cite{yolov5} & None & Modified CSP v5~\cite{yolov5} & 89.70 & 36.00 & 56.20 & 47.80 & 22.80 & 42.90 & 51.60  \\ 
CFP$_{\textrm{YOLOv5-M}}$~\cite{yolov5} & None & Modified CSP v5~\cite{yolov5} & 85.90 & 43.20 & 62.90 & 48.50 & 29.10 & 49.40 & 53.30  \\ 
CFP$_{\textrm{YOLOv5-L}}$~\cite{yolov5} & None & Modified CSP v5~\cite{yolov5} & \textbf{69.70} & \textbf{46.60} & \textbf{64.90} & \textbf{50.00} & \textbf{30.40} & \textbf{51.70} & \textbf{59.50}  \\ 
\cdashline{1-9}[0.8pt/2pt]
CFP$_{\textrm{YOLOX}}$~\cite{ge2021yolox} & None & DarkNet-53~\cite{redmon2018yolov3} & 85.30 & 47.00 & 66.90 & 52.90 & 31.70 & 53.10 & 61.60  \\
CFP$_{\textrm{YOLOX-S}}$~\cite{ge2021yolox} & None & Modified CSP v5~\cite{yolov5} & 87.10 & 41.10 & 60.00 & 44.50 & 24.20 & 45.40 & 54.50  \\
CFP$_{\textrm{YOLOX-M}}$~\cite{ge2021yolox} & None & Modified CSP v5~\cite{yolov5} & 76.40 & 46.40 & 65.10 & 50.30 & 29.40 & 51.20 & 60.50  \\
CFP$_{\textrm{YOLOX-L}}$~\cite{ge2021yolox} & None & Modified CSP v5~\cite{yolov5} & \textbf{66.30} & \textbf{49.40} & \textbf{67.80} & \textbf{53.60} & \textbf{32.40} & \textbf{54.30} & \textbf{64.00}  
\end{tabular}
\label{tab7}}
\end{center}
\end{table*}

\myparagraph{Result comparisons on speed and accuracy.} 
We perform a series of comparisons on the MS-COCO \emph{val} set with single-stage as well as two-stage detectors, and the results are shown in Table~\ref{tab7}. 
We can first see the two-stage object detection models, including the Faster R-CNN series with different backbone networks, Mask R-CNN, and D2Det. Our CFP$_{\textrm{YOLOX-L}}$ model has significant advantages in terms of precision, as well as inference speed and time.
Immediately after, we divide the single-stage detection methods into three parts in chronological order and then analyze them. 
There is no doubt that the proposed CFP$_{\textrm{YOLOX-L}}$ method improves the mAP by up to $27.80$\% compared to YOLOv3-ultralytics and its previous detectors. With nearly the same average precision, the CFP$_{\textrm{YOLOv5-M}}$ inferred 1.5 times faster compared to the EfficientDet-D2 detector. And comparing CFP$_{\textrm{YOLOX-L}}$ with EfficientDet-D3, not only the average accuracy is improved by $1.9$\%, but also the inference speed is $1.8$ times higher. In addition, in the comparison with YOLOv4 series, it can be found that the mAP of CFP$_{\textrm{YOLOv5-L}}$ is improved by $2.7$\% compared to YOLOv4-CSP. Besides, we can see all scaled YOLOv5 models, including YOLOv5-S, YOLOv5-M, and YOLOv5-L. The average precision of its best YOLOv5-L model is $1.4$\% lower than the CFP$_{\textrm{YOLOv5-L}}$. In the same way, our CFP method obtains a maximum average accuracy of $49.40$\%, which is $1.6$\% higher than YOLOX-L. 

\myparagraph{Qualitative Results on MS-COCO $2017$ \emph{test} set.} 
In addition, we also show in Figure~\ref{fig6} some visualization results of baseline (YOLOX-L), EVC$_{\textrm{YOLOX-L}}$ and CFP$_{\textrm{YOLOX-L}}$ on MS-COCOCO \emph{test} set. 
It is worth noting that we use white, red and orange boxes to mark where the detection task failures respectively.
White boxes indicate misses due to occlusion, light influence, or small object size. Red boxes indicate detection errors due to insufficient contextual semantic relationships, e.g., causing one object to be detected as two objects. The yellow boxes indicate an error in the object classification.

As can be seen in the first line of the figure, the detection result of YOLOX-L in the part marked in the white box is not ideal due to the distance factor of ``zebra''. And the EVC$_{\textrm{YOLOX-L}}$ can partially detect the ``zebra'' at a distance. Therefore, it is intuitively proved that EVC is very effective for small object detection in some intensive detection tasks.
In the second line of the figure, YOLOX-L does not fully detect the ``Cups'' in the cabinet due to factors such as occlusion and illumination. The EVC$_{\textrm{YOLOX-L}}$ model alleviates this problem by using MLP structures to capture the long-range dependencies of the features in the object. Finally, the CFP$_{\textrm{YOLOX-L}}$ model uses the GCR-assisted EVC scheme and gets better results.
In the third line of the figure, the CFP$_{\textrm{YOLOX-L}}$ model performs better in complex scenarios. Based on the EVC scheme, GCR is used to adjust intra-layer features for top-downm, and CFP$_{\textrm{YOLOX-L}}$ can solve the problem of classification better.
\section{Conclusions and Future Work}
In this work, we proposed a CFP for object detection, which was based on a globally explicit centralized feature regulation. We first proposed a spatial explicit visual center scheme, where a lightweight MLP was used to capture the globally long-range dependencies and a parallel learnable visual center was used to capture the local corner regions of the input images. Based on the proposed EVC, we then proposed a GCR for a feature pyramid in a top-down manner, where the explicit visual center information obtained from the deepest intra-layer feature was used to regulate all frontal shallow features. Compared to the existing methods, CFP not only has the ability to capture the global long-range dependencies, but also efficiently obtain an all-round yet discriminative feature representation. Experimental results on MS-COCO dataset verified that our CFP can achieve the consistent performance gains on the state-of-the-art object detection baselines. 
CFP is a generalized approach that can not only extract global long-range dependencies of the intra-layer features but also preserve the local corner regional information as much as possible, which is very important for dense prediction tasks. Therefore, in the future, we will start to develop some advanced intra-layer feature regulate methods to further improve the feature representation ability. Besides, we will try to apply EVC and GCR to other feature pyramid-based computer vision tasks, \eg, semantic segmentation, object localization, instance segmentation and person re-identification.
\bibliographystyle{IEEEtran}
\bibliography{IEEE_ref}

\begin{thebibliography}{10}
\providecommand{\url}[1]{#1}
\csname url@samestyle\endcsname
\providecommand{\newblock}{\relax}
\providecommand{\bibinfo}[2]{#2}
\providecommand{\BIBentrySTDinterwordspacing}{\spaceskip=0pt\relax}
\providecommand{\BIBentryALTinterwordstretchfactor}{4}
\providecommand{\BIBentryALTinterwordspacing}{\spaceskip=\fontdimen2\font plus
\BIBentryALTinterwordstretchfactor\fontdimen3\font minus
  \fontdimen4\font\relax}
\providecommand{\BIBforeignlanguage}[2]{{%
\expandafter\ifx\csname l@#1\endcsname\relax
\typeout{** WARNING: IEEEtran.bst: No hyphenation pattern has been}%
\typeout{** loaded for the language `#1'. Using the pattern for}%
\typeout{** the default language instead.}%
\else
\language=\csname l@#1\endcsname
\fi
#2}}
\providecommand{\BIBdecl}{\relax}
\BIBdecl

\bibitem{zhao2019object}
Z.-Q. Zhao, P.~Zheng, S.-t. Xu, and X.~Wu, ``Object detection with deep
  learning: A review,'' \emph{IEEE Transactions on Neural Networks and Learning
  Systems}, vol.~30, no.~11, pp. 3212--3232, 2019.

\bibitem{treml2016speeding}
M.~Treml, J.~Arjona-Medina, T.~Unterthiner, R.~Durgesh, F.~Friedmann,
  P.~Schuberth, A.~Mayr, M.~Heusel, M.~Hofmarcher, M.~Widrich \emph{et~al.},
  ``Speeding up semantic segmentation for autonomous driving,'' in \emph{Neural
  Information Processing Systems (NeurIPS)}, 2016.

\bibitem{havaei2017brain}
M.~Havaei, A.~Davy, D.~Warde-Farley, A.~Biard, A.~Courville, Y.~Bengio, C.~Pal,
  P.-M. Jodoin, and H.~Larochelle, ``Brain tumor segmentation with deep neural
  networks,'' \emph{Medical Image Analysis}, vol.~35, pp. 18--31, 2017.

\bibitem{girshick2015fast}
R.~Girshick, ``Fast r-cnn,'' in \emph{International Conference on Computer
  Vision (ICCV)}, 2015.

\bibitem{ren2015faster}
S.~Ren, K.~He, R.~Girshick, and J.~Sun, ``Faster r-cnn: Towards real-time
  object detection with region proposal networks,'' \emph{IEEE Transactions on
  Pattern Analysis and Machine Intelligence}, vol.~39, no.~6, pp. 1137--1149,
  2017.

\bibitem{liu2016ssd}
W.~Liu, D.~Anguelov, D.~Erhan, C.~Szegedy, S.~Reed, C.-Y. Fu, and A.~C. Berg,
  ``Ssd: Single shot multibox detector,'' in \emph{European Conference on
  Computer Vision (ECCV)}, 2016.

\bibitem{redmon2016you}
J.~Redmon, S.~Divvala, R.~Girshick, and A.~Farhadi, ``You only look once:
  Unified, real-time object detection,'' in \emph{IEEE Conference on Computer
  Vision and Pattern Recognition (CVPR)}, 2016.

\bibitem{dollar2014fast}
P.~Doll{\'a}r, R.~Appel, S.~Belongie, and P.~Perona, ``Fast feature pyramids
  for object detection,'' \emph{IEEE Transactions on Pattern Analysis and
  Machine Intelligence}, vol.~36, no.~8, pp. 1532--1545, 2014.

\bibitem{vashishth2020interacte}
S.~Vashishth, S.~Sanyal, V.~Nitin, N.~Agrawal, and P.~Talukdar, ``Interacte:
  Improving convolution-based knowledge graph embeddings by increasing feature
  interactions,'' in \emph{AAAI conference on Artificial Intelligence (AAAI)},
  2020.

\bibitem{zhang2018context}
H.~Zhang, K.~Dana, J.~Shi, Z.~Zhang, X.~Wang, A.~Tyagi, and A.~Agrawal,
  ``Context encoding for semantic segmentation,'' in \emph{IEEE Conference on
  Computer Vision and Pattern Recognition (CVPR)}, 2018.

\bibitem{zhang2019co}
H.~Zhang, H.~Zhang, C.~Wang, and J.~Xie, ``Co-occurrent features in semantic
  segmentation,'' in \emph{IEEE Conference on Computer Vision and Pattern
  Recognition (CVPR)}, 2019.

\bibitem{tan2020efficientdet}
M.~Tan, R.~Pang, and Q.~V. Le, ``Efficientdet: Scalable and efficient object
  detection,'' in \emph{IEEE Conference on Computer Vision and Pattern
  Recognition (CVPR)}, 2020.

\bibitem{ghiasi2019fpn}
G.~Ghiasi, T.-Y. Lin, and Q.~V. Le, ``Nas-fpn: Learning scalable feature
  pyramid architecture for object detection,'' in \emph{IEEE Conference on
  Computer Vision and Pattern Recognition (CVPR)}, 2019.

\bibitem{chen2020feature}
K.~Chen, Y.~Cao, C.~C. Loy, D.~Lin, and C.~Feichtenhofer, ``Feature pyramid
  grids,'' \emph{arXiv}, 2020.

\bibitem{zhang2020feature}
D.~Zhang, H.~Zhang, J.~Tang, M.~Wang, X.~Hua, and Q.~Sun, ``Feature pyramid
  transformer,'' in \emph{European Conference on Computer Vision (ECCV)}, 2020.

\bibitem{liu2018path}
S.~Liu, L.~Qi, H.~Qin, J.~Shi, and J.~Jia, ``Path aggregation network for
  instance segmentation,'' in \emph{IEEE Conference on Computer Vision and
  Pattern Recognition (CVPR)}, 2018.

\bibitem{lin2017feature}
T.-Y. Lin, P.~Doll{\'a}r, R.~Girshick, K.~He, B.~Hariharan, and S.~Belongie,
  ``Feature pyramid networks for object detection,'' in \emph{IEEE Conference
  on Computer Vision and Pattern Recognition (CVPR)}, 2017.

\bibitem{yin2020disentangled}
M.~Yin, Z.~Yao, Y.~Cao, X.~Li, Z.~Zhang, S.~Lin, and H.~Hu, ``Disentangled
  non-local neural networks,'' in \emph{European Conference on Computer Vision
  (ECCV)}, 2020.

\bibitem{wang2018non}
X.~Wang, R.~Girshick, A.~Gupta, and K.~He, ``Non-local neural networks,'' in
  \emph{IEEE Conference on Computer Vision and Pattern Recognition (CVPR)},
  2018.

\bibitem{vaswani2017attention}
A.~Vaswani, N.~Shazeer, N.~Parmar, J.~Uszkoreit, L.~Jones, A.~N. Gomez,
  {\L}.~Kaiser, and I.~Polosukhin, ``Attention is all you need,'' in
  \emph{Neural Information Processing Systems (NeurIPS)}, 2017.

\bibitem{luo2017non}
Z.~Luo, A.~Mishra, A.~Achkar, J.~Eichel, S.~Li, and P.-M. Jodoin, ``Non-local
  deep features for salient object detection,'' in \emph{IEEE Conference on
  Computer Vision and Pattern Recognition (CVPR)}, 2017.

\bibitem{cao2019gcnet}
Y.~Cao, J.~Xu, S.~Lin, F.~Wei, and H.~Hu, ``Gcnet: Non-local networks meet
  squeeze-excitation networks and beyond,'' in \emph{International Conference
  on Computer Vision Workshops (ICCVW)}, 2019.

\bibitem{carion2020end}
N.~Carion, F.~Massa, G.~Synnaeve, N.~Usunier, A.~Kirillov, and S.~Zagoruyko,
  ``End-to-end object detection with transformers,'' in \emph{European
  Conference on Computer Vision (ECCV)}, 2020.

\bibitem{dosovitskiy2020image}
A.~Dosovitskiy, L.~Beyer, A.~Kolesnikov, D.~Weissenborn, X.~Zhai,
  T.~Unterthiner, M.~Dehghani, M.~Minderer, G.~Heigold, S.~Gelly \emph{et~al.},
  ``An image is worth 16x16 words: Transformers for image recognition at
  scale,'' \emph{arXiv}, 2020.

\bibitem{liu2021swin}
Z.~Liu, Y.~Lin, Y.~Cao, H.~Hu, Y.~Wei, Z.~Zhang, S.~Lin, and B.~Guo, ``Swin
  transformer: Hierarchical vision transformer using shifted windows,'' in
  \emph{International Conference on Computer Vision (ICCV)}, 2021.

\bibitem{wang2021pyramid}
W.~Wang, E.~Xie, X.~Li, D.-P. Fan, K.~Song, D.~Liang, T.~Lu, P.~Luo, and
  L.~Shao, ``Pyramid vision transformer: A versatile backbone for dense
  prediction without convolutions,'' in \emph{IEEE Conference on Computer
  Vision and Pattern Recognition (CVPR)}, 2021.

\bibitem{radosavovic2020designing}
I.~Radosavovic, R.~P. Kosaraju, R.~Girshick, K.~He, and P.~Doll{\'a}r,
  ``Designing network design spaces,'' in \emph{IEEE Conference on Computer
  Vision and Pattern Recognition (CVPR)}, 2020.

\bibitem{zhou2016learning}
B.~Zhou, A.~Khosla, A.~Lapedriza, A.~Oliva, and A.~Torralba, ``Learning deep
  features for discriminative localization,'' in \emph{IEEE Conference on
  Computer Vision and Pattern Recognition (CVPR)}, 2016.

\bibitem{ru2022learning}
L.~Ru, Y.~Zhan, B.~Yu, and B.~Du, ``Learning affinity from attention:
  End-to-end weakly-supervised semantic segmentation with transformers,''
  \emph{arXiv}, 2022.

\bibitem{li2022transcam}
R.~Li, Z.~Mai, C.~Trabelsi, Z.~Zhang, J.~Jang, and S.~Sanner, ``Transcam:
  Transformer attention-based cam refinement for weakly supervised semantic
  segmentation,'' \emph{arXiv}, 2022.

\bibitem{strudel2021segmenter}
R.~Strudel, R.~Garcia, I.~Laptev, and C.~Schmid, ``Segmenter: Transformer for
  semantic segmentation,'' in \emph{International Conference on Computer Vision
  (ICCV)}, 2021.

\bibitem{zhu2021unified}
F.~Zhu, Y.~Zhu, L.~Zhang, C.~Wu, Y.~Fu, and M.~Li, ``A unified efficient
  pyramid transformer for semantic segmentation,'' in \emph{International
  Conference on Computer Vision Workshops (ICCVW)}, 2021.

\bibitem{zhang2021self}
D.~Zhang, H.~Zhang, J.~Tang, X.-S. Hua, and Q.~Sun, ``Self-regulation for
  semantic segmentation,'' in \emph{International Conference on Computer Vision
  (ICCV)}, 2021.

\bibitem{lin2014microsoft}
T.-Y. Lin, M.~Maire, S.~Belongie, J.~Hays, P.~Perona, D.~Ramanan,
  P.~Doll{\'a}r, and C.~L. Zitnick, ``Microsoft coco: Common objects in
  context,'' in \emph{European Conference on Computer Vision (ECCV)}, 2014.

\bibitem{yolov5}
\BIBentryALTinterwordspacing
{YOLOv5}. [Online]. Available: \url{{https://github.com/ultralytics/yolov5}}
\BIBentrySTDinterwordspacing

\bibitem{ge2021yolox}
Z.~Ge, S.~Liu, F.~Wang, Z.~Li, and J.~Sun, ``Yolox: Exceeding yolo series in
  2021,'' \emph{arXiv}, 2021.

\bibitem{zhao2019m2det}
Q.~Zhao, T.~Sheng, Y.~Wang, Z.~Tang, Y.~Chen, L.~Cai, and H.~Ling, ``M2det: A
  single-shot object detector based on multi-level feature pyramid network,''
  in \emph{AAAI Conference on Artificial Intelligence (AAAI)}, 2019.

\bibitem{krizhevsky2012imagenet}
A.~Krizhevsky, I.~Sutskever, and G.~E. Hinton, ``Imagenet classification with
  deep convolutional neural networks,'' \emph{Neural Information Processing
  Systems (NeurIPS)}, vol.~25, no.~6, pp. 80--90, 2017.

\bibitem{touvron2021training}
H.~Touvron, M.~Cord, M.~Douze, F.~Massa, A.~Sablayrolles, and H.~J{\'e}gou,
  ``Training data-efficient image transformers \& distillation through
  attention,'' in \emph{International Conference on Machine Learning (ICML)},
  2021.

\bibitem{zhu2020deformable}
X.~Zhu, W.~Su, L.~Lu, B.~Li, X.~Wang, and J.~Dai, ``Deformable detr: Deformable
  transformers for end-to-end object detection,'' \emph{arXiv}, 2020.

\bibitem{beal2020toward}
J.~Beal, E.~Kim, E.~Tzeng, D.~H. Park, A.~Zhai, and D.~Kislyuk, ``Toward
  transformer-based object detection,'' \emph{arXiv}, 2020.

\bibitem{zheng2021rethinking}
S.~Zheng, J.~Lu, H.~Zhao, X.~Zhu, Z.~Luo, Y.~Wang, Y.~Fu, J.~Feng, T.~Xiang,
  P.~H. Torr \emph{et~al.}, ``Rethinking semantic segmentation from a
  sequence-to-sequence perspective with transformers,'' in \emph{IEEE
  Conference on Computer Vision and Pattern Recognition (CVPR)}, 2021.

\bibitem{chen20182}
Y.~Chen, Y.~Kalantidis, J.~Li, S.~Yan, and J.~Feng, ``A\^{} 2-nets: Double
  attention networks,'' in \emph{Neural Information Processing Systems
  (NeurIPS)}, 2018.

\bibitem{ramachandran2019stand}
P.~Ramachandran, N.~Parmar, A.~Vaswani, I.~Bello, A.~Levskaya, and J.~Shlens,
  ``Stand-alone self-attention in vision models,'' in \emph{Neural Information
  Processing Systems (NeurIPS)}, 2019.

\bibitem{vaswani2021scaling}
A.~Vaswani, P.~Ramachandran, A.~Srinivas, N.~Parmar, B.~Hechtman, and
  J.~Shlens, ``Scaling local self-attention for parameter efficient visual
  backbones,'' in \emph{IEEE Conference on Computer Vision and Pattern
  Recognition (CVPR)}, 2021.

\bibitem{tolstikhin2021mlp}
I.~Tolstikhin, N.~Houlsby, A.~Kolesnikov, L.~Beyer, X.~Zhai, T.~Unterthiner,
  J.~Yung, D.~Keysers, J.~Uszkoreit, M.~Lucic \emph{et~al.}, ``Mlp-mixer: An
  all-mlp architecture for vision,'' in \emph{Neural Information Processing
  Systems (NeurIPS)}, 2021.

\bibitem{liu2021pay}
H.~Liu, Z.~Dai, D.~R. So, and Q.~V. Le, ``Pay attention to mlps,'' in
  \emph{Neural Information Processing Systems (NeurIPS)}, 2021.

\bibitem{yu2021metaformer}
W.~Yu, M.~Luo, P.~Zhou, C.~Si, Y.~Zhou, X.~Wang, J.~Feng, and S.~Yan,
  ``Metaformer is actually what you need for vision,'' \emph{arXiv}, 2021.

\bibitem{hou2022vision}
Q.~Hou, Z.~Jiang, L.~Yuan, M.-M. Cheng, S.~Yan, and J.~Feng, ``Vision
  permutator: A permutable mlp-like architecture for visual recognition,''
  \emph{arXiv}, 2022.

\bibitem{girshick2014rich}
R.~Girshick, J.~Donahue, T.~Darrell, and J.~Malik, ``Rich feature hierarchies
  for accurate object detection and semantic segmentation,'' in \emph{IEEE
  Conference on Computer Vision and Pattern Recognition (CVPR)}, 2014.

\bibitem{dai2016r}
J.~Dai, Y.~Li, K.~He, and J.~Sun, ``R-fcn: Object detection via region-based
  fully convolutional networks,'' in \emph{Neural Information Processing
  Systems (NeurIPS)}, 2016.

\bibitem{he2017mask}
K.~He, G.~Gkioxari, P.~Doll{\'a}r, and R.~Girshick, ``Mask r-cnn,'' in
  \emph{International Conference on Computer Vision (ICCV)}, 2017.

\bibitem{redmon2018yolov3}
J.~Redmon and A.~Farhadi, ``Yolov3: An incremental improvement,'' \emph{arXiv},
  2018.

\bibitem{bochkovskiy2020yolov4}
A.~Bochkovskiy, C.-Y. Wang, and H.-Y.~M. Liao, ``Yolov4: Optimal speed and
  accuracy of object detection,'' \emph{arXiv}, 2020.

\bibitem{peng2021conformer}
Z.~Peng, W.~Huang, S.~Gu, L.~Xie, Y.~Wang, J.~Jiao, and Q.~Ye, ``Conformer:
  Local features coupling global representations for visual recognition,'' in
  \emph{International Conference on Computer Vision (ICCV)}, 2021.

\bibitem{wang2020cspnet}
C.-Y. Wang, H.-Y.~M. Liao, Y.-H. Wu, P.-Y. Chen, J.-W. Hsieh, and I.-H. Yeh,
  ``Cspnet: A new backbone that can enhance learning capability of cnn,'' in
  \emph{IEEE Conference on Computer Vision and Pattern Recognition Workshops
  (CVPRW)}, 2020.

\bibitem{howard2017mobilenets}
A.~G. Howard, M.~Zhu, B.~Chen, D.~Kalenichenko, W.~Wang, T.~Weyand,
  M.~Andreetto, and H.~Adam, ``Mobilenets: Efficient convolutional neural
  networks for mobile vision applications,'' \emph{arXiv}, 2017.

\bibitem{larsson2016fractalnet}
G.~Larsson, M.~Maire, and G.~Shakhnarovich, ``Fractalnet: Ultra-deep neural
  networks without residuals,'' \emph{arXiv}, 2016.

\bibitem{liu2019simple}
J.-J. Liu, Q.~Hou, M.-M. Cheng, J.~Feng, and J.~Jiang, ``A simple pooling-based
  design for real-time salient object detection,'' in \emph{IEEE Conference on
  Computer Vision and Pattern Recognition (CVPR)}, 2019.

\bibitem{zhang2017Mixup}
H.~Zhang, M.~Cisse, Y.~N. Dauphin, and D.~Lopez-Paz, ``mixup: Beyond empirical
  risk minimization,'' in \emph{International Conference on Learning
  Representations (ICLR)}, 2018.

\bibitem{ramachandran2017swish}
P.~Ramachandran, B.~Zoph, and Q.~V. Le, ``Swish: a self-gated activation
  function,'' \emph{arXiv}, 2017.

\bibitem{he2015spatial}
K.~He, X.~Zhang, S.~Ren, and J.~Sun, ``Spatial pyramid pooling in deep
  convolutional networks for visual recognition,'' \emph{IEEE Transactions on
  Pattern Analysis and Machine Intelligence}, vol.~37, no.~9, pp. 1904--1916,
  2015.

\bibitem{he2016deep}
S.~R. Kaiming~He, Xiangyu~Zhang and J.~Sun, ``Deep residual learning for image
  recognition,'' in \emph{IEEE Conference on Computer Vision and Pattern
  Recognition (CVPR)}, 2016.

\bibitem{goyal2017accurate}
P.~Goyal, P.~Doll{\'a}r, R.~Girshick, P.~Noordhuis, L.~Wesolowski, A.~Kyrola,
  A.~Tulloch, Y.~Jia, and K.~He, ``Accurate, large minibatch sgd: Training
  imagenet in 1 hour,'' \emph{arXiv}, 2017.

\bibitem{lian2021mlp}
D.~Lian, Z.~Yu, X.~Sun, and S.~Gao, ``As-mlp: An axial shifted mlp architecture
  for vision,'' \emph{arXiv}, 2021.

\bibitem{chen2021recurrent}
Z.~Chen, J.~Zhang, and D.~Tao, ``Recurrent glimpse-based decoder for detection
  with transformer,'' \emph{arXiv}, 2021.

\bibitem{xie2017aggregated}
S.~Xie, R.~Girshick, P.~Doll{\'a}r, Z.~Tu, and K.~He, ``Aggregated residual
  transformations for deep neural networks,'' in \emph{IEEE Conference on
  Computer Vision and Pattern Recognition (CVPR)}, 2017.

\bibitem{fang2021you}
Y.~Fang, B.~Liao, X.~Wang, J.~Fang, J.~Qi, R.~Wu, J.~Niu, and W.~Liu, ``You
  only look at one sequence: Rethinking transformer in vision through object
  detection,'' in \emph{Neural Information Processing Systems (NeurIPS)}, 2021.

\bibitem{song2021vidt}
H.~Song, D.~Sun, S.~Chun, V.~Jampani, D.~Han, B.~Heo, W.~Kim, and M.-H. Yang,
  ``Vidt: An efficient and effective fully transformer-based object detector,''
  \emph{arXiv}, 2021.

\bibitem{cao2020d2det}
J.~Cao, H.~Cholakkal, R.~M. Anwer, F.~S. Khan, Y.~Pang, and L.~Shao, ``D2det:
  Towards high quality object detection and instance segmentation,'' in
  \emph{IEEE Conference on Computer Vision and Pattern Recognition (CVPR)},
  2020.

\bibitem{redmon2017yolo9000}
J.~Redmon and A.~Farhadi, ``Yolo9000: better, faster, stronger,'' in \emph{IEEE
  Conference on Computer Vision and Pattern Recognition (CVPR)}, 2017.

\bibitem{fu2017dssd}
C.-Y. Fu, W.~Liu, A.~Ranga, A.~Tyagi, and A.~C. Berg, ``Dssd: Deconvolutional
  single shot detector,'' \emph{arXiv}, 2017.

\bibitem{Tan_2020_CVPR}
M.~Tan, R.~Pang, and Q.~V. Le, ``Efficientdet: Scalable and efficient object
  detection,'' in \emph{IEEE Conference on Computer Vision and Pattern
  Recognition (CVPR)}, 2020.

\bibitem{tan2019efficientnet}
M.~Tan and Q.~Le, ``Efficientnet: Rethinking model scaling for convolutional
  neural networks,'' in \emph{International Conference on Machine Learning
  (ICML)}, 2019.

\bibitem{wang2021scaled}
C.-Y. Wang, A.~Bochkovskiy, and H.-Y.~M. Liao, ``Scaled-yolov4: Scaling cross
  stage partial network,'' in \emph{IEEE Conference on Computer Vision and
  Pattern Recognition (CVPR)}, 2021.

\end{thebibliography}
\end{document}